\pdfoutput=1

\documentclass[11pt]{article}

\usepackage[]{EMNLP2023}
\usepackage{color}

\usepackage{times}
\usepackage{latexsym}

\usepackage[T1]{fontenc}

\usepackage[utf8]{inputenc}

\usepackage{microtype}

\usepackage{inconsolata}

\usepackage{hyperref}
\usepackage{color}
\usepackage{url}
\usepackage{adjustbox}
\usepackage{makecell}
\usepackage{tabularx}
\usepackage[T1]{fontenc}
\usepackage{xcolor}
\usepackage{booktabs}
\usepackage{multirow}
\usepackage[shortlabels]{enumitem}

\usepackage{algorithm}
\usepackage{algpseudocode}

\newcommand{\gptt}{\textsc{GPT-3}}

\newcommand{\llama}{\textsc{LLaMA}}
\newcommand{\llamaseven}{\textsc{LLaMA-7B}}

\newcommand{\chatgpt}{\textsc{ChatGPT}}
\newcommand{\claimLLM}{\textsc{Examinee}}
\newcommand{\examinerLLM}{\textsc{Examiner}}
\newcommand{\crossexamination}{\textsc{LMvLM}}

\newcommand{\nl}[1]{\textit{``#1''}}

\definecolor{light_blue}{HTML}{e1f6ff}
\usepackage[most]{tcolorbox}
\tcbset{on line, 
        boxsep=1pt, left=0pt,right=0pt,top=0pt,bottom=0pt,
        colframe=white,
        colback=light_blue,  
        highlight math style={enhanced}
        }

\title{LM vs LM: Detecting Factual Errors via Cross Examination}

\author{Roi Cohen$^1$~~~~May Hamri$^1$~~~~Mor Geva$^2$~~~~Amir Globerson$^{1,3}$ \vspace{3pt}\\
$^1$Blavatnik School of Computer Science, Tel Aviv University\vspace{3pt}\\$^2$Google DeepMind~~~$^3$Google Research\\
\small{\texttt{\{roi1, mayhamri\}@mail.tau.ac.il, pipek@google.com, gamir@tauex.tau.ac.il}}\\
}

\begin{document}
\maketitle

\begin{abstract}
A prominent weakness of modern language models (LMs) is their tendency to generate factually incorrect text, which hinders their usability. A natural question is whether such factual errors can be detected automatically. Inspired by truth-seeking mechanisms in law, we propose a  factuality evaluation framework for LMs that is based on cross-examination.
Our key idea is that an incorrect claim is likely to result in inconsistency with other claims that the model generates. To discover such inconsistencies, we facilitate a multi-turn interaction between the LM that generated the claim and another LM (acting as an examiner) which introduces questions to discover inconsistencies. 
We empirically evaluate our method on factual claims made by multiple recent LMs on four benchmarks, finding that it
outperforms existing methods and baselines, often by a large gap.
Our results demonstrate the potential of using interacting LMs to capture factual errors.

\end{abstract}

\section{Introduction}

\begin{figure}[t]
\setlength{\belowcaptionskip}{-10pt}
    \centering    \includegraphics[width=\linewidth]{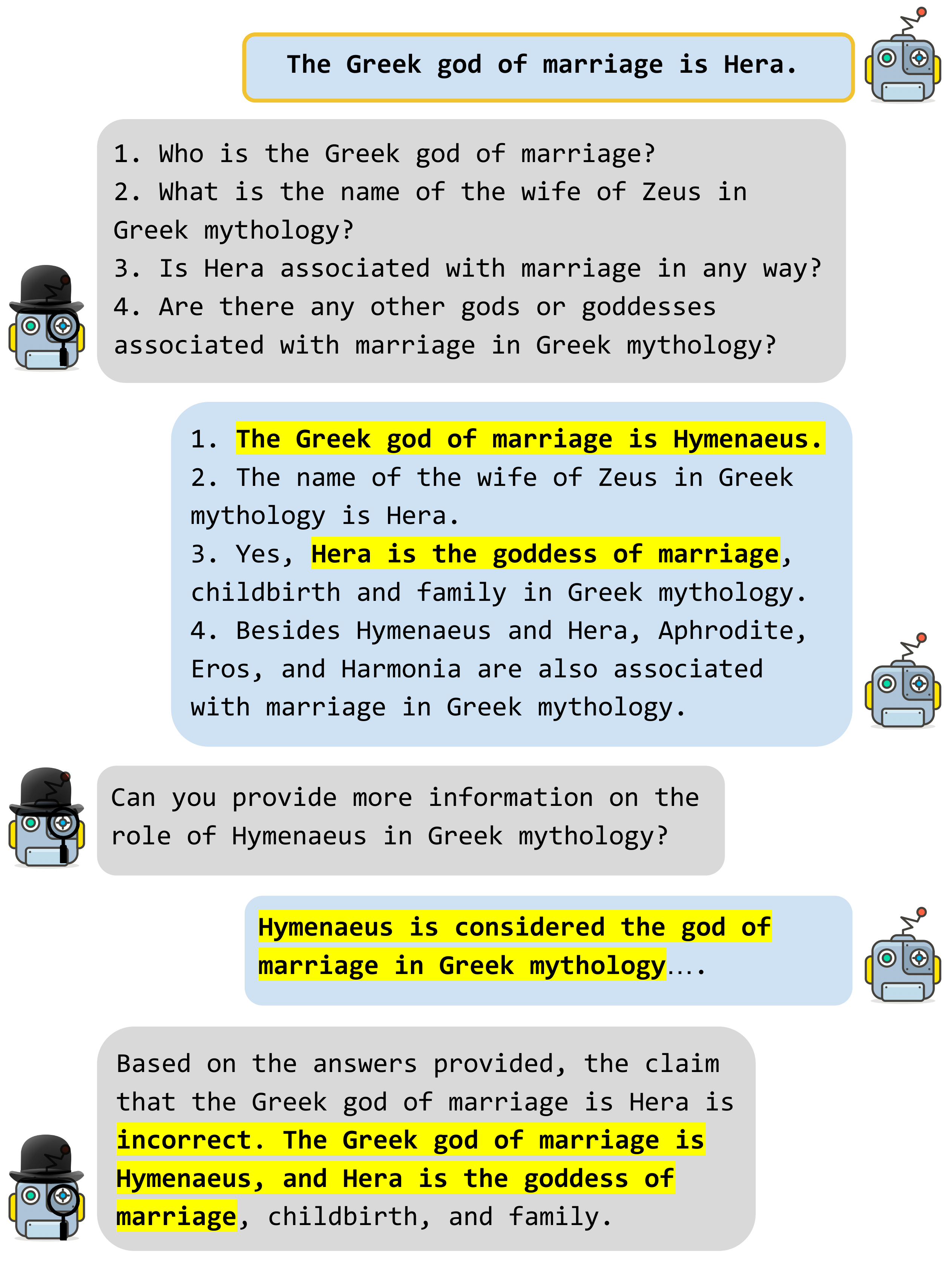}
    \caption{An example of our \crossexamination{} approach. The first line shows the statement made by the  \claimLLM{} LLM. Then an interaction between the \examinerLLM{} and \claimLLM{} takes place, and the \examinerLLM{} arrives at a conclusion whether the original statement was correct or not (here it concludes that it was a false statement).}
    \label{figure:intro}
\end{figure}

Modern language models (LMs) often generate inconsistent \cite{elazar-etal-2021-measuring}, non-attributable \cite{rashkin2021measuring, bohnet2022attributed, liu2023evaluating}, or factually incorrect text \cite{tam2022evaluating, devaraj-etal-2022-evaluating, maynez-etal-2020-faithfulness}, thus negatively impacting the reliability of these models \cite{amodei2016concrete, hendrycks2021unsolved}.
This has prompted the community to develop methods that calibrate the confidence of model predictions to better align with their quality \cite{brundage2020toward}. For example, prior methods have used probabilistic approaches \cite{jiang-etal-2020-know, zablotskaia2023uncertainty} clustering \cite{kuhn2023semantic}, fine-tuning \cite{kadavath2022language, lin2022teaching} and in-context learning \cite{alivanistos2022prompting, cohen-etal-2023-crawling}.

In this work, we take a different approach to this problem, motivated by truth-seeking mechanisms in law. Specifically, we consider the setting where a witness is cross-examined in order to check whether their statement is factually correct or not. In such a setting, the examiner asks questions that aim to lead towards contradictory statements by the witness, while a contradiction implies that the witness lied at least in some of the statements, hence the well known quote \textit{``Were you lying then or are you lying now?''} \cite{wilder1957witness}. 

To employ this mechanism to LM factual calibration, we propose the following setting, illustrated in Figure~\ref{figure:intro}. Our goal is to check whether a statement made by an LM (\nl{The Greek god of marriage is Hera}) is factually correct. We refer to the model that generated this statement as the \claimLLM{}. To check whether this fact is correct, we use another LM, called \examinerLLM{}, to conduct a cross-examination of \claimLLM{}. Concretely, we craft designated prompts to facilitate a multi-turn interaction between the two LMs, where \examinerLLM{} issues questions (e.g., \nl{Is Hera associated with marriage in any way?}) to \claimLLM{} to check the veracity of the original statement. The examination is concluded by a decision from \examinerLLM{} as to whether the original claim was correct or not.\footnote{In practice, \examinerLLM{} and \claimLLM{} can be the same LM (e.g., \gptt{}) that is prompted in two different ways to define its different roles.}

Our problem setting is related to that of calibration \cite{pmlr-v70-guo17a}, where the goal is to predict the probability at which a model will err.
However, unlike previous approaches to this problem, we use  text generated by LMs. 
Our approach is motivated by the intuition that calibration is actually an elaborate reasoning process where one checks the level of support that a fact has based on other statements the model believes. We argue that such complex reasoning is naturally performed via the strong conversational skills of modern LMs. 

We use our method to detect errors in LM generation in the context of factual question-answering. Our experiments with several recent LMs -- \chatgpt{}, \gptt{} \cite{brown2020language, ouyang2022training}, and \llama{} \cite{touvron2023llama} -- show that cross-examination effectively detects factually incorrect claims generated by LMs. Specifically, across multiple datasets and examination settings, it detects over 70\% of the incorrect claims while maintaining a high precision of $>$80\%, outperforming strong baselines by a large gap.

Further analysis shows that examiner LMs introduce multiple questions throughout the examination, and employ various strategies to reveal inconsistencies, including question paraphrasing, validation of implicated arguments, claim decomposition, and requests for evidence.

To conclude, our contributions are (a) framing the task of factuality testing as an interaction between two LMs, (b) proposing a concrete implementation of this interaction via the use of one LM with different prompts in a zero-shot setting, and (c) demonstrating improved factuality detection accuracy across several benchmarks.

\section{LM Cross-Examination}

Our goal is to employ an ``examiner'' LM (\examinerLLM{}) to evaluate claims generated by another LM (\claimLLM{}). To this end, we leverage the recent success of prompting \cite{liu2023pre}, to facilitate a cross-examination setting between the two LMs.
In such a setting, \examinerLLM{} should introduce questions with the objective of revealing inconsistencies with respect to an initial claim made by \claimLLM{}. Such inconsistencies can be considered as a signal for uncertainty of \claimLLM{} in its original claim, and thus, can be used to assess whether its original statement was correct.

\begin{figure}[t]
\setlength{\belowcaptionskip}{-10pt}
    \centering
    \includegraphics[scale=0.5]{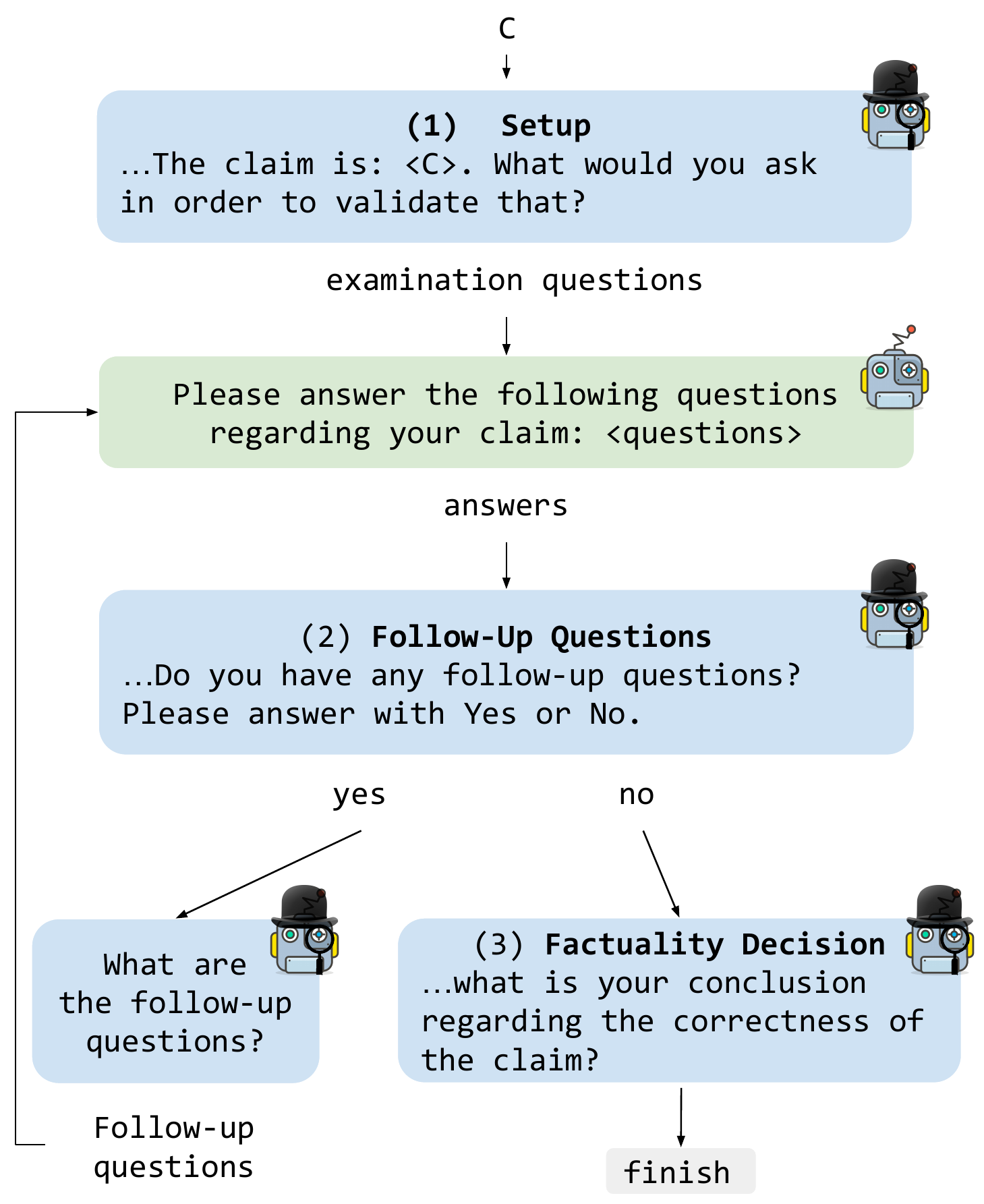}
    \caption{The three-stage  process of cross-examination between the \examinerLLM{} and \claimLLM{}, where the factuality of a claim $C$ generated by \claimLLM{} is estimated by \examinerLLM{}.}
    \label{figure:examination_process_diagram}
\end{figure}

Given an \examinerLLM{} LM and a claim $C$ generated by an \claimLLM{}, our method establishes a multi-turn interaction between the LMs, where at each turn the other LM is prompted with a designated prompt that incorporates the outputs from previous turns. This interaction continues until the examiner has no further questions and can provide its final decision.
To establish a meaningful interaction that reveals possible inconsistencies, we define three stages for the examination, each guided by a specific prompt. As part of each prompt for \claimLLM{} or \examinerLLM{}, we provide the outputs generated in the previous rounds for context. We next describe the examination stages in detail, with the overall process illustrated in Figure~\ref{figure:examination_process_diagram}.

\paragraph{Stage 1: Setup}
The examination begins by ``assigning'' the \examinerLLM{} its role. Namely, describing the task setting, providing it with the \claimLLM{}'s claim, and asking it to generate questions for the \claimLLM{}.\footnote{We observe that this effectively steers \examinerLLM{} to ask natural questions directly related to the given claim $C$ (\S\ref{sec:analysis}).}

Next, we feed the questions generated by \examinerLLM{}, one at a time, to \claimLLM{}, concatenated to the following instructions: \texttt{Please answer the following questions regarding your claim}. The response from \claimLLM{} yields a set of answers to the questions from \examinerLLM{}.

\paragraph{Stage 2: Follow-up Questions}
We next feed \examinerLLM{} with the answers generated by \claimLLM{} to its initial questions, and ask \examinerLLM{} whether it has any follow-up questions.
Notably, outputs from \examinerLLM{} at this stage are conditioned on the previous output from \claimLLM{}.
If the answer from \examinerLLM{} is ``Yes'', we then further prompt it to obtain more questions. 
This phase is conducted iteratively, until either \examinerLLM{} declares it has no follow-up questions, or the number of turns has reached a threshold.\footnote{We use a maximum of five turns in our experiments.}

\paragraph{Stage 3: Factuality Decision}
Once no further questions are obtained from \examinerLLM{}, we prompt it to conclude whether the claim $C$ is true or false.
Specifically, we request it to reply with either ``correct'' or ``incorrect'' as its final conclusion. 
In cases where the examiner does not output either of ``correct'' or ``incorrect'', we consider its final decision to be a rejection of the claim. 
Typically though, we observe that the examiner follows the instructions and indeed generates a definitive conclusion (see statistics in \S\ref{sec:analysis}).

\section{Related Work}

\paragraph{Attribution and Fact Checking} 
Our goal is closely related to works on attribution and fact verification. Namely, checking if a LM-generated text is faithful to some source text \cite{bohnet2022attributed, honovich2022true}. This problem has been addressed via several approaches, including question generation \cite{wang-etal-2020-asking, honovich-etal-2021-q2, scialom-etal-2021-questeval}, NLI \cite{thorne-etal-2018-fact, welleck-etal-2019-dialogue, maynez-etal-2020-faithfulness, dziri-etal-2022-evaluating, gao2022rarr, kamoi-etal-2023-shortcomings}, data augmentation \cite{atanasova-etal-2022-fact, wright-etal-2022-generating, gekhman2023trueteacher}, and planning schemes that allow the model to self-edit its own generation \cite{schick2022peer}.
Unlike these works, we are not assuming any reference text or an external knowledge base.
Instead, we directly check if the LM's claim is likely to be correct, by probing 
the model for inconsistencies. Our approach also uses multi-turn dialogue as a key component.

\paragraph{Model Calibration}
A key challenge with prediction models is to provide a probability of the answer being incorrect, a problem known as model calibration \cite{pmlr-v70-guo17a}. The problem of factual-error detection can be viewed as a variation of calibration, where instead of a continuous probability, we provide a binary prediction for whether the model is correct or not. This is also related to the setting of selective prediction, where a model can choose to abstain from answering a query \cite{varshney-etal-2022-investigating, kamath2020selective}.
Common approaches to calibration are to perform various transformations on model logits \cite{desai2020calibration, jiang-etal-2021-know}, and measuring uncertainty \citep[e.g., see][]{kuhn2023semantic}. 
More recent works have studied the use of LMs for providing calibration, by training them on statements known to be factually correct or incorrect. This ``supervised'' approach has been explored via fine-tuning \cite{kadavath2022language, lin2022teaching} and in-context learning \cite{cohen-etal-2023-crawling, alivanistos2022prompting}.

Our work focuses on zero-shot factual error detection
that involves just two categories: predicting whether a model's claim is correct or incorrect. We propose a novel approach to this problem, using multi-turn LLM interaction.
While we focus on a binary setting, one could envision an extension of our approach to continuous outputs (for example, to output a probabilistic estimation for the correctness of the claim).

\paragraph{Multi-Agent LMs}
Using multiple LMs in an interactive manner is a relatively new idea with many potential applications. It has been shown that LMs can utilize additional LMs or tools to better solve downstream tasks \cite{schick2023toolformer}. Additionally, \citet{ParkPCMLB22} showed that in a social setting, LMs demonstrate certain social skills that emerge from this interaction, and \citet{reflexion} proposes that a LM can use a different model to instruct it when to ``reflect'' on its recent action, while performing a planned sequence of actions aimed at solving a given query. Intuitively, this model detects signs of hallucination or inefficient planning within the LM's trajectory. 


\paragraph{Consistency Across Generations}
LMs have been shown to generate inconsistent outputs given different prompt paraphrases \cite{elazar-etal-2021-measuring, newman2021p}. Prior work showed that prompts can be automatically optimized to produce factually correct claims more robustly \cite{lester2021power, zhong2021factual, qin-eisner-2021-learning}. \citet{hao2022bertnet} utilized multiple generated paraphrases to gauge consistency \cite{hao2022bertnet}, and other works \cite{elazar-etal-2021-measuring, zhou2022prompt} further proposed training objectives to improve model consistency. Another approach to handling multiple outputs is via variants of decoding strategies \cite{wang2022self}, or model ensembles \cite{sun2022quantifying}.
In our work, we build on these, assuming inconsistencies are more likely to occur with incorrect claims, and let an examiner model search for these by introducing questions to the examinee.

\paragraph{Chain of Thought Reasoning}
Recent work has shown that LMs can be prompted to elaborate on their reasoning process, to self-ask themselves follow-up questions, before reaching a final conclusion, and that this could be exploited to improve mathematical, multi-hop and common-sense reasoning skills \cite{wei2022chain, press2022measuring, yoran2023answering}, along with planning and problem-solving abilities \cite{huang2022inner, long2023large}.
Another interesting approach to complex reasoning in LMs is recent work on Maieutic prompting \cite{Jung2022MaieuticPL}, that answers a question by recursively generating a set of facts and reasoning over those.

Our approach may be viewed as constructing an elaborate chain-of-thought explanation for the examinee's claim. However, we do not train this explanation via in-context or fine-tuning, and rather rely on different prompts for its generation.

\begin{table*}[ht]
\setlength{\belowcaptionskip}{-10pt}
\setlength\tabcolsep{4.0pt}
\footnotesize
\begin{center}
\begin{tabular}{ p{0.12\linewidth} p{0.25\linewidth} p{0.56\linewidth}}
\textbf{Stage} & \textbf{GPT3 Prompt(s)} & \textbf{ChatGPT Prompt(s)} \\ 
\toprule
(1) Setup
& \texttt{Imagine trying to prove that a claim that someone claims is true, is wrong. 
You have the opportunity to ask any question in order to prove that the claim is wrong.
The claim is: <$C$>.}

\texttt{What would you ask in order to validate that?}

& \texttt{Your goal is to try to verify the correctness of the following claim:<$C$>, based on the background information you will gather. To gather this, You will provide short questions whose purpose will be to verify the correctness of the claim, and I will reply to you with the answers to these. Hopefully, with the help of the background questions and their answers, you will be able to reach a conclusion as to whether the claim is correct or possibly incorrect. Please keep asking questions as long as you're yet to be sure regarding the true veracity of the claim. Please start with the first questions.}
 \\
\midrule
(2) Follow-Up Questions 
& 
\multicolumn{2}{p{0.8\linewidth}}{(i)~~\texttt{Do you have any follow-up questions? Please answer with Yes or No.}

(ii) \texttt{What are the follow-up questions?}} \\
\midrule
(3) Factuality Decision 
& \multicolumn{2}{p{0.8\linewidth}}{\texttt{Based on the interviewee's answers to your questions, what is your conclusion regarding the correctness of the claim? Do you think it is correct or incorrect?}} \\
\bottomrule
\end{tabular}
\end{center}
\caption{Prompts provided to \examinerLLM{} in each stage of the examination, with respect to a claim $C$ by \claimLLM{}.
}
\label{table:prompts}
\end{table*}

\section{Experiments}
\label{sec:experiments}

In this section, we conduct experiments on multiple datasets and models to evaluate our approach, focusing on the task of factual question-answering.

\subsection{Experimental Setup}

\paragraph{Factual Question Answering}
\label{claim_generation_setup}
One key use-case of LMs is answering questions seeking factual knowledge. For example, \nl{How old was Barack Obama when he was first elected?}. In such cases, it is crucial for the model to answer the question correctly, or to indicate that it does not know the answer.
We thus evaluate our approach on several Question Answering and Fact Completion datasets. These are typically provided as a set of $(Q ,A)$ pairs of a question $Q$ and its ground-truth answer $A$. 
Having gold answers allows us to evaluate if a predicted answer is factually correct or not, which can be used to evaluate our \crossexamination{} approach. 

To apply cross-examination in this setting, we first convert the answer predicted by the model into a \claimLLM{} claim that can be provided as input to the examination procedure. Formally, given a question $Q$, if $Q$ is phrased as a fill-in-the-blank question (e.g. \nl{Bailey Peninsula is located in \textunderscore\textunderscore\textunderscore\textunderscore}), then we feed it to the \claimLLM{} model  to obtain a prediction that completes the sentence and forms a claim. In cases where $Q$ is phrased as a question (e.g. \nl{Where is Bailey Peninsula located?}), we prompt the model to provide an answer in a claim format with: \texttt{``Please answer the following question: <$Q$> Please phrase your answer as a claim.''}
This process results in a claim $C$ that states the model's ``belief'' about the answer to $Q$. We then evaluate the truthfulness of $C$ through cross-examination, and compare the examiner's decision of whether $C$ is correct or not to the ground-truth correctness.

\begin{table}[t]
\setlength{\belowcaptionskip}{-10pt}
    \footnotesize
    \centering
    \begin{tabular}{lcccc}
         \claimLLM{} & LAMA & TriviaQA & NQ & PopQA   \\
         \midrule
         \llamaseven{} &53.9 &48.4 &33.8 &24.9 \\
         \gptt{} &79.8 &74.2 &50.1 &43.9 \\
         \chatgpt{} &80.9 &77.2 &53.3 &45.6 \\
         \bottomrule
    \end{tabular}
    \caption{Portion of factually correct claims by every \claimLLM{} LM on each dataset.}
    \label{table:examinee_accuracy}
\end{table}

\paragraph{Factuality Evaluation Labels}
To evaluate our method, it is necessary to have ``gold decisions'' to compare the examiner's decisions against. Such labels can be obtained from the ground-truth answers in the data, namely, the decision for a claim $C$ is correct if it matches an evaluation of $C$ against the gold answer $A$.
To evaluate if the claim $C$ obtained for a question $Q$ is correct with respect to the ground-truth answer $A$, we first check if $A$ or any of its aliases (if provided as part of the dataset, e.g., ``FC Tottenham'' and ``Tottenham Hotspur'') appears as a sub-string in $C$ \cite{schick2023toolformer, meng2022locating}. Next, to avoid incorrect labels resulting from this automatic evaluation \cite{bulian-etal-2022-tomayto}, we manually review all the claims marked as incorrect in the first step, and fix any labeling mistakes. We also filter out any ambiguous or unclear claims generated by \claimLLM{}.

\paragraph{Examiner Evaluation} 
We evaluate how well the examiner detects claims that are factually incorrect, using the following metrics:\footnote{We say that the examiner ``rejects'' a claim if the examiner concludes that the claim is incorrect.} 
\begin{itemize}
[leftmargin=*,topsep=2pt,itemsep=2pt,parsep=0pt]

    \item \textbf{Precision}: the portion of incorrect claims, out of the claims rejected by the examiner.

    \item \textbf{Recall}: 
    the portion of incorrect claims rejected by the examiner, out of all the incorrect claims.

    \item \textbf{F1}: the harmonic mean of precision and recall.
\end{itemize}
For completeness, we additionally report (in \S\ref{apx:example_examinations}) the complementary Precision, Recall, and F1 scores with respect to detection of correct claims.

\paragraph{Data}
\label{datasets_setup}
We consider the following datasets: LAMA \cite{petroni2019language}, TriviaQA \cite{joshi2017triviaqa}, Natural Questions (NQ) \cite{kwiatkowski2019natural} and PopQA \cite{mallen2022not}. These datasets cover a wide range of queries, from real user queries (NQ), to trivia questions (TriviaQA), and subject-relation-object facts phrased as queries (LAMA, PopQA). We consider the closed-book open-ended setting, where we do not provide any context or answer choices to the model. 
We evaluate our approach on 1,000 random examples from the test set (or from the development set if a test set is not available).\footnote{We use only a subset of examples due to the high cost of executing large LMs like \gptt{}, which we include in our experiments.}

In addition, we created a dataset of false claims to further test our approach. This ``Falsehoods'' dataset contains only wrong claims, created separately for each model (\gptt{} and \chatgpt{}) and for each of the four QA datasets. Concretely, given a model and a question $Q$, we prompt the model to generate a false answer (see \S\ref{apx:fake_facts} for details). We verify that these are indeed incorrect claims by checking that the gold answer (and any of its aliases, if they exist) does not occur in the generated text. This yields a subset of examples that are realistic, namely, the answer matches the target type (e.g., ``a city'') but is incorrect (see examples in Table~\ref{table:false_claims}). The examiner's decision for these examples should always be to reject.

\begin{table}[t]
\setlength{\belowcaptionskip}{-10pt}
    \footnotesize
    \centering
    \setlength{\tabcolsep}{2.8pt}
    \begin{tabular}{lp{3.7cm}p{3.2cm}}
         & \textbf{False claim} & \textbf{Correct claim}  \\ \midrule
        \parbox[t]{2mm}{\multirow{8}{*}{\rotatebox[origin=c]{90}{\gptt{}}}} & \nl{Wanlockhead is the highest village in France because it is located in the French Alps.} & \nl{Wanlockhead is the highest village in Scotland, a country in Europe.} \\ \cmidrule{2-3}
        & \nl{Louis Oosthuizen the 2010 Open Golf Champion is American, because he was born in the United States.} & \nl{Louis Oosthuizen, the 2010 Open Golf Champion, is South African.} \\ 
         \midrule
        \parbox[t]{2mm}{\multirow{6}{*}{\rotatebox[origin=c]{90}{\chatgpt{}}}} & \nl{The screenwriter for "Smile" was definitely Steven Spielberg.} & \nl{The screenwriter for "Smile" was Jerry Belson.} \\ \cmidrule{2-3}
        & \nl{Fontenay is located in the beautiful country of Antarctica.} & \nl{Fontenay is located in France.} \\
         \bottomrule
    \end{tabular}
    \caption{Example false claims generated by \chatgpt{} for PopQA and by \gptt{} for TriviaQA.}
    \label{table:false_claims}
\end{table}

\begin{table*}[t]
\setlength\tabcolsep{4.0pt}
\footnotesize
\begin{center}
\begin{tabular}{l  ccc | ccc | ccc | ccc }
 & \multicolumn{3}{c}{LAMA} & \multicolumn{3}{c}{TriviaQA} &  \multicolumn{3}{c}{NQ} & \multicolumn{3}{c}{PopQA} \\ [0.1cm]
\multicolumn{1}{c}{} & \textbf{P}  & \textbf{R} & \textbf{F1}  & \textbf{P} & \textbf{R} & \textbf{F1}  & \textbf{P}  & \textbf{R} & \textbf{F1} & \textbf{P}  & \textbf{R} & \textbf{F1}  \\
\toprule

AYS                 & $82.3$    &$25.2$   &$38.6$   
                    & $79.9$    &$17.9$   &$29.2$                    
                    & $85.2$    &$29.1$   &$43.3$ 
                    & $78.4$    &$35.7$   &$63.9$ \\
IDK                 & $49.1$    &$52.4$   &$50.7$
                    & $48.7$    &$66.5$   &$56.2$
                    & $62.5$    &$60.7$   &$61.6$
                    & $70.0$    &$61.1$   &$65.2    $ \\ \midrule
\textbf{\crossexamination{}}   & $85.1$    &$70.7$   &$76.7$
                            & $82.8$    &$71.6$   &$76.8$
                            & $74.5$    &$74.9$   &$77.7$ 
                            & $83.6$    &$77.1$   &$80.2$  \\
\textbf{\crossexamination{}} (Majority)   & $\mathbf{86.6}$    &$\mathbf{75.8}$   &$\mathbf{80.8}$
                            & $\mathbf{84.5}$    &$\mathbf{80.8}$   &$\mathbf{82.6}$
                            & $\mathbf{82.3}$    &$\mathbf{76.1}$   &$\mathbf{79.1}$ 
                            & $\mathbf{87.0}$    &$\mathbf{84.0}$   &$\mathbf{85.4}$  \\ 
~~~~ - Follow-up  & $83.8$    &$68.1$   &$75.1$ 
                            & $82.3$    &$69.7$   &$75.5$ 
                            & $74.8$    &$72.1$   &$73.4$ 
                            & $82.0$    &$73.3$   &$77.4$   \\ 

\bottomrule
\end{tabular}
\end{center}
\caption{Precision (P), Recall (R), and F1 scores for \crossexamination{} with \chatgpt{} as \examinerLLM{} and \claimLLM{}, compared to baselines. The last row shows an ablation of our method without the follow-up questions stage.}
\label{table:results_chatgpt_chatgpt}
\end{table*}

\begin{table*}[t]
\setlength{\belowcaptionskip}{-10pt}
\setlength\tabcolsep{4.0pt}
\footnotesize
\begin{center}
\begin{tabular}{l  ccc | ccc | ccc | ccc }
 & \multicolumn{3}{c}{LAMA} & \multicolumn{3}{c}{TriviaQA} &  \multicolumn{3}{c}{NQ} & \multicolumn{3}{c}{PopQA} \\ [0.1cm]
\multicolumn{1}{c}{} & \textbf{P}  & \textbf{R} & \textbf{F1}  & \textbf{P} & \textbf{R} & \textbf{F1}  & \textbf{P}  & \textbf{R} & \textbf{F1} & \textbf{P}  & \textbf{R} & \textbf{F1}  \\
\toprule
AYS                 & $74.8$    &$17.9$   &$28.9$  
                    & $80.3$    &$19.8$   &$31.8$ 
                    & $74.9$    &$20.7$   &$32.3$ 
                    & $74.6$    &$22.7$   &$34.8$   \\
IDK                 & $43.0$    &$42.1$   &$42.5$ 
                    & $47.9$    &$45.7$   &$46.7$ 
                    & $60.9$    &$45.3$   &$52.0$ 
                    & $52.1$    &$37.6$   &$43.7$     \\
Confidence-Based    & $38.6$    &$\mathbf{85.8}$   &$53.2$ 
                    & $39.6$    &$\mathbf{84.4}$   &$53.9$ 
                    & $56.2$    &$72.7$   &$63.4$ 
                    & $60.8$    &$69.7$   &$64.9$   \\ 
IC-IDK             & $71.5$    &$46.3$   &$56.2$ 
                    & $70.6$    &$49.7$   &$60.1$ 
                    & $70.0$    &$57.6$   &$63.2$  
                    & $76.9$    &$37.7$   &$50.6$   \\  \midrule
\textbf{\crossexamination{}}   & $78.8$    &$69.9$   &$74.1$
                            & $81.6$    &$64.6$   &$72.1$
                            & $70.5$    &$66.6$   &$68.5$ 
                            & $75.5$    &$69.1$   &$72.2$  \\
\textbf{\crossexamination{}} (Majority)   & $\mathbf{80.7}$    &$77.9$   &$\mathbf{79.3}$ 
                            & $\mathbf{83.1}$    &$72.1$   &$\mathbf{77.2}$ 
                            & $\mathbf{79.3}$    &$\mathbf{76.8}$   &$\mathbf{78.0}$ 
                            & $\mathbf{82.2}$    &$\mathbf{71.4}$   &$\mathbf{76.4}$   \\[0.05cm]
~~~~ - Follow-up  & $76.4$    &$71.1$   &$73.7$ 
                            & $78.7$    &$64.8$   &$71.1$ 
                            & $66.6$    &$70.1$   &$68.3$ 
                            & $70.9$    &$65.8$   &$68.3$   \\ 
\bottomrule
\end{tabular}
\end{center}
\caption{Precision (P), Recall (R), and F1 scores for \crossexamination{} with \gptt{} as \examinerLLM{} and \claimLLM{}, compared to baselines. The last row shows an ablation of our method without the follow-up questions stage.}
\label{table:results_gpt_gpt}
\end{table*}

\paragraph{Models}
We use \chatgpt{} (\texttt{gpt-3.5-turbo}), \gptt{} (\texttt{text-davinci-003}) \cite{brown2020language, ouyang2022training}, and \llama{}-7B \cite{touvron2023llama}, in three \examinerLLM{} vs. \claimLLM{} cross-examination settings: \gptt{} vs. \gptt{}, \chatgpt{} vs. \chatgpt{}, and \chatgpt{} vs. \llama{}. Notably, using the same LM as \examinerLLM{} and \claimLLM{} (except for their prompts, which are different), provides a cleaner setting where both LMs share the same knowledge. The prompts used for each LM at every stage of the examination are shown in Table~\ref{table:prompts}.

\begin{table*}[t]
\setlength{\belowcaptionskip}{-10pt}
\setlength\tabcolsep{4.0pt}
\footnotesize
\begin{center}
\begin{tabular}{l  lll | lll | lll | lll }
 & \multicolumn{3}{c}{LAMA} & \multicolumn{3}{c}{TriviaQA} &  \multicolumn{3}{c}{NQ} & \multicolumn{3}{c}{PopQA} \\ [0.1cm]
\multicolumn{1}{c}{} & \textbf{P}  & \textbf{R} & \textbf{F1}  & \textbf{P} & \textbf{R} & \textbf{F1}  & \textbf{P}  & \textbf{R} & \textbf{F1} & \textbf{P}  & \textbf{R} & \textbf{F1}  \\
\toprule
AYS                 & $61.4$    &$38.0$   &$46.9$  
                    & $60.0$    &$35.7$   &$44.8$ 
                    & $71.1$    &$15.0$   &$24.8$ 
                    & $74.8$    &$14.2$   &$23.9$  \\
IC-IDK             & $56.6$    &$49.0$   &$52.5$ 
                    & $58.9$    &$52.5$   &$55.5$ 
                    & $66.2$    &$53.4$   &$59.1$ 
                    & $66.8$    &$50.1$   &$57.3$   \\
IDK                 & $61.6$    &$44.8$   &$51.9$ 
                    & $62.0$    &$32.9$   &$43.0$ 
                    & $64.4$    &$12.1$   &$20.4$ 
                    & $66.7$    &$16.8$   &$26.8$    \\ 
Confidence-Based    & $54.9$    &$\mathbf{76.7}$   &$64.0$ 
                    & $56.9$    &$\mathbf{85.8}$   &$68.4$ 
                    & $64.4$    &$63.5$ &$63.9$ 
                    & $64.6$    &$53.6$   &$58.6$  \\ \midrule
\textbf{\crossexamination{}}   & $81.1$    &$66.4$   &$73.0$
                            & $80.1$    &$70.8$   &$75.2$
                            & $79.3$    &$65.5$   &$71.7$ 
                            & $84.9$    &$73.6$   &$78.8$  \\
\textbf{\crossexamination{}} (Majority)   & $\mathbf{82.9}$    &$73.9$   &$\mathbf{78.1}$  
                            & $\mathbf{80.3}$    &$76.8$  &$\mathbf{78.5}$  
                            & $\mathbf{83.7}$    &$\mathbf{74.2}$   &$\mathbf{78.7}$ 
                            & $\mathbf{88.3}$  & $\mathbf{77.4}$  &$\mathbf{82.5}$  \\
~~~~ - Follow-up  & $79.7$    &$65.7$   &$72.0$ 
                            & $80.0$    &$69.8$   &$74.6$ 
                            & $79.4$    &$63.7$   &$70.7$ 
                            & $83.3$    &$71.8$   &$77.1$   \\ 
\bottomrule
\end{tabular}
\end{center}
\caption{Precision (P), Recall (R), and F1 scores for \crossexamination{} with \chatgpt{} as \examinerLLM{} and \llama{} as \claimLLM{}, compared to baselines. The last row is an ablation of our method without the follow-up questions stage.}
\label{table:results_llama_chatgpt}
\end{table*}

\paragraph{Baselines}
For each setting, we compare \crossexamination{} with recent methods for uncertainty detection and variants of our approach:
\begin{itemize}
[leftmargin=*,topsep=2pt,itemsep=2pt,parsep=0pt]
    \item \textbf{Confidence-Based}:
    The prediction head of LMs outputs a probability for the predicted token. It is a common practice to use this probability as a measure of confidence in the prediction \cite{yoshikawa-okazaki-2023-selective}. In our case, the LM generates a multi-token claim, and we calculate the confidence for the claim as the product of probabilities for all predicted tokens of the answer only.
    In order to output a binary decision (i.e., is the claim correct or not), we optimize a threshold over the train dataset to maximize F1. Note that our examination approach does not require tuning any threshold.
    
    \item \textbf{Are you sure? (AYS)}: Recent work \cite{kadavath2022language,cohen-etal-2023-crawling} has shown that LMs can be trained to estimate their certainty in generated facts. Here, we use a zero-shot version of this approach where we directly ``ask'' the model whether it is sure. Specifically, we add the following prompt right after the claim generation: \texttt{``Are you sure regarding the correctness of your claim? Please answer with Yes or No''}. Then we take the output as the prediction whether the claim is correct or not.

    \item \textbf{I don't know (IDK)}: Recently, \citet{ganguli2023capacity} showed that LMs might have the capability to self-correct themselves, when instructed to do so. Here we instruct the model to output \emph{``I don't know''} if it is uncertain, by concatenating the following sentence to the original query: \texttt{"If you are not sure you know the answer, answer with 'I don't know' only."}. 
    If the model answers `I don't know' we label the corresponding claim as false, and otherwise true.

    \item \textbf{In-context IDK (IC-IDK)}: We teach the model to output that it doesn't know the answer, via in-context demonstrations. We follow \citet{cohen-etal-2023-crawling} and test each of the queries within an in-context setting. For each query, we first provide the model with $K$ demonstrations, as $D$ of them labeled as \emph{``Don't know''} examples, while the rest $K-D$ are provided with their gold answer from the dataset. The \emph{``Don't know''} examples are randomly selected out of a set of examples the model failed on, while evaluating it on an held-out set of examples from the dataset in a zero-shot setting. Intuitively, these examples' answers are likely to be unknown to the model, hence we labeled them with \emph{``Don't know''}. The model predictions are either a target text or \emph{``Don't know''}. Based on the output, we generate a factuality label as in the IDK baseline above. Notably, this baseline requires labeled data for the in-context demonstrations, which is not necessary for our approach.

    \item \textbf{\crossexamination{}}: A single execution of our method, where we accept or reject the claim according to the examiner's final decision.
    
    \item \textbf{\crossexamination{} (Majority)}:
    For a given claim, we apply our method three times (with the same \examinerLLM{} and \claimLLM{}), using sampling generation for follow-up questions generation. We reject the claim in case at least two of the examinations concluded it is false.

\end{itemize}

Since output probabilities are not provided as part of the \chatgpt{}'s API, we cannot provide results for the Confidence-Based baselines in this case. Moreover, we observe that \chatgpt{} often fails to understand the task of IC-IDK.

\begin{table}[t]
\setlength{\belowcaptionskip}{-10pt}
\setlength\tabcolsep{3.0pt}
\footnotesize
\begin{center}
\begin{tabular}{lcccc}
 & LAMA & TriviaQA & NQ & PopQA \\ 
\midrule
\gptt{}    & 65.7  &98.4  &89.9  &83.1  \\ 
\gptt{} (Majority)    &75.8   &98.5   &92.0   &88.0   \\ \midrule
\chatgpt{}    & 83.6  & 97.9  & 90.4  & 88.8  \\
\chatgpt{} (Majority)    &87.1   &98.6   &94.2   &93.9   \\
\bottomrule
\end{tabular}
\end{center}
\caption{Accuracy of \gptt{} and \chatgpt{} as \examinerLLM{} on false claims generated for each dataset.}
\label{table:false_claims_accuracy}
\end{table}

\subsection{Results}
Tables~\ref{table:results_chatgpt_chatgpt},~\ref{table:results_gpt_gpt},~\ref{table:results_llama_chatgpt} show the results for the settings \chatgpt{} vs. \chatgpt{}, \gptt{} vs. \gptt{}, and \llama{} vs. \chatgpt{}, respectively.
Across all settings, our method outperforms the baselines, often by a large gap. For example, it obtains $85.4$ F1 compared to $\leq65.2$ by baselines for \chatgpt{} on PopQA (Table \ref{table:results_chatgpt_chatgpt}), and $77.2$ F1 compared to $\leq60.1$ for \gptt{} on TriviaQA (Table \ref{table:results_gpt_gpt}).
Notably, the most substantial gains are in terms of recall, showing the superiority of our method in detecting factually incorrect claims (when compared to the baselines which achieve reasonable precision too). 
Interestingly, we observe that \chatgpt{} generally outperforms \gptt{}.

Last, Table~\ref{table:false_claims_accuracy} shows the accuracy of our method and baselines on our Falsehood dataset. For both \chatgpt{} and \gptt{}, \crossexamination{} successfully rejects the vast majority of the false claims, obtaining 87\%-98\% accuracy with \chatgpt{} and 75\%-99\% with \gptt{} across all datasets.

\begin{table}[t]
\setlength{\belowcaptionskip}{-10pt}
\setlength\tabcolsep{3.0pt}
    \centering
    \footnotesize
    \begin{tabular}{p{2.4cm}p{1.6cm}p{1.2cm}p{1.6cm}}
         &  \chatgpt{} / & \gptt{} / & \chatgpt{} / \\
         &  \chatgpt{} & \gptt{} & \llama{} \\ \midrule
         $\#$ of questions &$7.0 \pm 2.8$ &$6.4 \pm 4.3$ &$6.8 \pm 4.4$ \\ \midrule
         $\#$ of follow-up questions per iteration &$1.3 \pm 1.0$ &$1.3 \pm 0.6$ &$1.1 \pm 0.5$ \\ \midrule
         $\#$ of follow-up iterations &$1.9 \pm 1.2$ &$1.3 \pm 0.7$ &$1.6 \pm 1.0$ \\ \midrule
         $\#$ of questions per iteration &$3.1 \pm 2.1$ &$2.7 \pm 1.6$ &$2.9 \pm 1.9$ \\ \midrule
         $\%$ of inconclusive examiner decisions &$14.8\%$ &$9.1\%$ &$10.3\%$ \\
         \bottomrule
    \end{tabular}
    \caption{Cross-examination statistics for each setting (\examinerLLM{}/\claimLLM{}), averaged across datasets.}
    \label{table:statistics}
\end{table}

\begin{table*}[t]
\setlength{\belowcaptionskip}{-10pt}
    \footnotesize
    \centering
    \begin{tabular}{p{2.2cm}p{12.8cm}}
        \textbf{Pattern} & \textbf{Example statements/questions generated by \examinerLLM{} during examination}  \\
        \midrule
        
        \multirow{2}{=}{Rephrasing the claim} & Claim: \nl{The first Fast and Furious film was released in 2001.} \\
        & \texttt{In which year was the first Fast and Furious film released?} \\
        \midrule
        
        \multirow{2}{=}{Rephrasing Questions} & Claim: \nl{The screenwriter who is credited with writing the screenplay for Winner is Wendy Riss} \\ & 1. \texttt{What is the name of the screenwriter who is credited with writing the screenplay for Winner?} 2. \texttt{Who is credited with writing the screenplay for Winner?} \\
        \midrule
        
        \multirow{2}{=}{Validation of Implications} & Claim: \nl{The director of The Town was Ben Affleck.} \\
        & \texttt{Is Ben Affleck known for directing any movies?}\\
        \midrule
        
        \multirow{2}{=}{Logical decomposition} & Claim: \nl{The second oldest of the Pevensie children in C S Lewis's The Lion, the Witch and the Wardrobe is Edmund.} \\
        & 1. \texttt{What is the birth order of the Pevensie children in C S Lewis's The Lion, the Witch and the Wardrobe?} 2. \texttt{What are their ages?} 3. \texttt{Who appears second in this list?}\\ \midrule
        
       \multirow{2}{=}{Request for attribution} & Claim: \nl{The screenwriter of Cover Up is Bill Blake} \\ & \texttt{Is there any evidence or documentation that supports the claim that Bill Blake was the screenwriter for Cover Up?} \\ \midrule
       
       \multirow{2}{=}{Wrong intermediate answers} & Claim: \nl{There are eight vertices (corners) on an octahedron.} \\ & \examinerLLM{}: \texttt{How many vertices does an octahedron have?} \claimLLM{}: \texttt{An octahedron has eight vertices, each of which is the point where three edges meet.} \\ 
        
         \bottomrule
    \end{tabular}
    \caption{Examples for frequent  patterns of \chatgpt{} and \gptt{} observed through manual analysis of cross-examinations.}
    \label{table:pattern_examples}
\end{table*}

\subsection{Ablations}
\label{subsec:ablations}

We perform an ablation, where we remove the follow-up iterations in the examination process to gauge their benefit. Results are reported for \gptt{} in Table \ref{table:results_gpt_gpt} (last row), showing a large decrease in performance (e.g. $78 \rightarrow 68.3$ in F1 for NQ and $77.2 \rightarrow 71.1$ for TriviaQA). Notably, recall scores are decreased by 6\%-10\%. Overall, this shows the importance of the follow-up questions issued by the examiner to assess the examinee's claim.

\section{Analysis of Cross-Examinations}
\label{sec:analysis}

We analyze cross-examinations by \gptt{} and \chatgpt{} to better understand the success and failure cases of our method. We find that examiner LMs typically ask multiple questions in the examination, and perhaps surprisingly, apply different strategies to reveal inconsistencies.

\paragraph{Examination Statistics}
Table \ref{table:statistics} provides statistics on the cross-examinations performed by \chatgpt{} and \gptt{}. Both models introduce multiple queries (6-7 on average) during an examination, with typically 1-2 steps of follow-up questions, which are important for the examiner's decision (\S\ref{subsec:ablations}).
We also observe a non-negligible number of claims (9\%-15\%) where the examiner LM does not arrive at a concrete final decision (i.e., it does not generate ``correct'' or ``incorrect'' as the final decision, we reject the claim in those cases). In our qualitative analysis, we identify reasons that could explain these cases.

\paragraph{Qualitative Analysis}
We manually analyze a sample of 96 examinations -- 48 by each LM, with 6 correct and 6 incorrect examinations for each model and each dataset. 
We observe the following trends (examples are in Table \ref{table:pattern_examples}):
\begin{enumerate}
[leftmargin=*,topsep=2pt,itemsep=2pt,parsep=0pt]
    \item \textbf{Rephrasing the claim}: In about $60\%$ of the examinations, both LMs introduce questions which are paraphrases of the original question. This supports the assumption that the \examinerLLM{} seeks  inconsistencies by generating variants of the original claim. 
    
    \item \textbf{Rephrasing Questions}: In about half of the examinations, both LMs introduce questions that are similar to previously asked questions or have a different phrasing. This is a desirable behavior as it can reveal inconsistencies if the examinee provides a different answer for the same question.  
    
    \item \textbf{Validation of logical implications}: The \examinerLLM{} asks \claimLLM{} regarding implied arguments that must be true whenever the original claim is correct. This can be observed in $70\%$ of the correct detections of \gptt{}, and $87.5\%$ out of the correct detections of \chatgpt{}.

    \item \textbf{Logical questions}: The \examinerLLM{} decomposes the claim into multiple sub-questions which together compose a trajectory to validating it. Such decompositions appear in about 75\% of the cases for \chatgpt{} but only 10\% in \gptt{}. We observe these in $33\%$ of the correct detections of \gptt{}, and $70\%$ for \chatgpt{}.

    \item \textbf{Request for attribution}: The \examinerLLM{} ask the \claimLLM{} about the existence of external evidence to support the claim. This happens in about 30\% of the cases for both LMs.

    \item \textbf{Wrong intermediate answers}: The \claimLLM{} responds with factually incorrect answers to one or more of the questions originated by the \examinerLLM{}. We observe this occurs mostly in cases where the original claim is false (it happens in only in about $14\%$ of the cases where the \claimLLM{} is correct). In both models, this can be observed in about half of the cases where the claim is false and  has also been detected by the \examinerLLM{}. Furthermore, it occurs in about $80\%$ of the cases where the \examinerLLM{} has accepted a false claim, and in $45\%$ where the \examinerLLM{} has rejected a correct claim.

\end{enumerate}

We note that in most cases where \crossexamination{} fails, \claimLLM{} has provided incorrect information to \examinerLLM{}. This might indicate that in those cases \claimLLM{} has encoded a large set of factually wrong facts that are mutually consistent, thus making it hard for the \examinerLLM{} to detect inconsistencies.
Finally, the fact that \chatgpt{} more commonly validates the claim through logical questions might be a key factor in its superiority over \gptt{} in our setting.

\section{Conclusion}
\label{sec:discussion}

We introduce \crossexamination{}, a method for zero-shot detection of factuality errors, inspired by the cross-examination practice employed in a court of law. Our method uses prompting to facilitate a multi-turn interaction between an examiner LM and an examinee LM, to reveal inconsistencies that imply factually incorrect claims. We evaluate \crossexamination{} in the context of factual question answering, showing it substantially improves detection of factual errors made by LMs.

Our method builds on a fundamental connection between self-consistency (i.e., consistency of an LM with itself) and factual consistency (i.e., consistency between factual claims generated by an LM and ground-truth facts). We consider the LM itself as the source of information, and we test whether a claim it has generated is faithful and consistent with several other beliefs it has. 

Our work can be extended in several ways.
First, \crossexamination{} provides interpretable information about related beliefs of the model, which could be analyzed to understand what makes the model commit certain mistakes.
Second, one may incorporate several LM instances into the factuality detection process, rather the having only a single \examinerLLM{}. Finally, one can train the \examinerLLM{} to generate questions more effectively.

\section*{Limitations}

We note three limitations of our method \crossexamination{}. First, unlike other methods, it requires multiple queries of the examinee and examiner LMs, which could be costly when using external APIs such as those used in this work. This could be a key consideration when scaling this approach to large numbers of claims.

Second, for our method to succeed, both LMs (\claimLLM{} and \examinerLLM{}), but mostly \examinerLLM{}, should be able to follow instructions and have the ability to reason over information in a relatively long context. This skill is currently mostly demonstrated by larger models (>10B parameters) and thus our method may not perform as well for smaller models.

Last, any logical flaws in the examiner's operation are likely to affect the overall examination, potentially leading to inaccurate decisions. However, our experiments show that, even if such flaws occur, our method is still useful on average as it substantially improves factuality detection.
Nonetheless, developing safety mechanisms that detect and mitigate logical flaws is an important research direction, that we leave for future work.


\section*{Acknowledgements}
We thank Roee Aharoni and Avi Caciularu for valuable feedback and constructive suggestions. This work is supported in part by the Israeli Science Foundation.

\bibliography{anthology,custom}

\begin{thebibliography}{65}
\expandafter\ifx\csname natexlab\endcsname\relax\def\natexlab#1{#1}\fi

\bibitem[{Alivanistos et~al.(2022)Alivanistos, Santamar{\'\i}a, Cochez, Kalo,
  van Krieken, and Thanapalasingam}]{alivanistos2022prompting}
Dimitrios Alivanistos, Selene~B{\'a}ez Santamar{\'\i}a, Michael Cochez,
  Jan-Christoph Kalo, Emile van Krieken, and Thiviyan Thanapalasingam. 2022.
\newblock Prompting as probing: Using language models for knowledge base
  construction.
\newblock \emph{arXiv preprint arXiv:2208.11057}.

\bibitem[{Amodei et~al.(2016)Amodei, Olah, Steinhardt, Christiano, Schulman,
  and Man{\'e}}]{amodei2016concrete}
Dario Amodei, Chris Olah, Jacob Steinhardt, Paul Christiano, John Schulman, and
  Dan Man{\'e}. 2016.
\newblock Concrete problems in ai safety.
\newblock \emph{arXiv preprint arXiv:1606.06565}.

\bibitem[{Atanasova et~al.(2022)Atanasova, Simonsen, Lioma, and
  Augenstein}]{atanasova-etal-2022-fact}
Pepa Atanasova, Jakob~Grue Simonsen, Christina Lioma, and Isabelle Augenstein.
  2022.
\newblock \href {https://doi.org/10.1162/tacl_a_00486} {Fact checking with
  insufficient evidence}.
\newblock \emph{Transactions of the Association for Computational Linguistics},
  10:746--763.

\bibitem[{Bohnet et~al.(2022)Bohnet, Tran, Verga, Aharoni, Andor, Soares,
  Eisenstein, Ganchev, Herzig, Hui et~al.}]{bohnet2022attributed}
Bernd Bohnet, Vinh~Q Tran, Pat Verga, Roee Aharoni, Daniel Andor, Livio~Baldini
  Soares, Jacob Eisenstein, Kuzman Ganchev, Jonathan Herzig, Kai Hui, et~al.
  2022.
\newblock Attributed question answering: Evaluation and modeling for attributed
  large language models.
\newblock \emph{arXiv preprint arXiv:2212.08037}.

\bibitem[{Brown et~al.(2020)Brown, Mann, Ryder, Subbiah, Kaplan, Dhariwal,
  Neelakantan, Shyam, Sastry, Askell et~al.}]{brown2020language}
Tom Brown, Benjamin Mann, Nick Ryder, Melanie Subbiah, Jared~D Kaplan, Prafulla
  Dhariwal, Arvind Neelakantan, Pranav Shyam, Girish Sastry, Amanda Askell,
  et~al. 2020.
\newblock Language models are few-shot learners.
\newblock \emph{Advances in neural information processing systems},
  33:1877--1901.

\bibitem[{Brundage et~al.(2020)Brundage, Avin, Wang, Belfield, Krueger,
  Hadfield, Khlaaf, Yang, Toner, Fong et~al.}]{brundage2020toward}
Miles Brundage, Shahar Avin, Jasmine Wang, Haydn Belfield, Gretchen Krueger,
  Gillian Hadfield, Heidy Khlaaf, Jingying Yang, Helen Toner, Ruth Fong, et~al.
  2020.
\newblock Toward trustworthy ai development: mechanisms for supporting
  verifiable claims.
\newblock \emph{arXiv preprint arXiv:2004.07213}.

\bibitem[{Bulian et~al.(2022)Bulian, Buck, Gajewski, B{\"o}rschinger, and
  Schuster}]{bulian-etal-2022-tomayto}
Jannis Bulian, Christian Buck, Wojciech Gajewski, Benjamin B{\"o}rschinger, and
  Tal Schuster. 2022.
\newblock \href {https://aclanthology.org/2022.emnlp-main.20} {Tomayto,
  tomahto. beyond token-level answer equivalence for question answering
  evaluation}.
\newblock In \emph{Proceedings of the 2022 Conference on Empirical Methods in
  Natural Language Processing}, pages 291--305, Abu Dhabi, United Arab
  Emirates. Association for Computational Linguistics.

\bibitem[{Cohen et~al.(2023)Cohen, Geva, Berant, and
  Globerson}]{cohen-etal-2023-crawling}
Roi Cohen, Mor Geva, Jonathan Berant, and Amir Globerson. 2023.
\newblock \href {https://aclanthology.org/2023.findings-eacl.139} {Crawling the
  internal knowledge-base of language models}.
\newblock In \emph{Findings of the Association for Computational Linguistics:
  EACL 2023}, pages 1856--1869, Dubrovnik, Croatia. Association for
  Computational Linguistics.

\bibitem[{Desai and Durrett(2020)}]{desai2020calibration}
Shrey Desai and Greg Durrett. 2020.
\newblock Calibration of pre-trained transformers.
\newblock \emph{arXiv preprint arXiv:2003.07892}.

\bibitem[{Devaraj et~al.(2022)Devaraj, Sheffield, Wallace, and
  Li}]{devaraj-etal-2022-evaluating}
Ashwin Devaraj, William Sheffield, Byron Wallace, and Junyi~Jessy Li. 2022.
\newblock \href {https://doi.org/10.18653/v1/2022.acl-long.506} {Evaluating
  factuality in text simplification}.
\newblock In \emph{Proceedings of the 60th Annual Meeting of the Association
  for Computational Linguistics (Volume 1: Long Papers)}, pages 7331--7345,
  Dublin, Ireland. Association for Computational Linguistics.

\bibitem[{Dziri et~al.(2022)Dziri, Rashkin, Linzen, and
  Reitter}]{dziri-etal-2022-evaluating}
Nouha Dziri, Hannah Rashkin, Tal Linzen, and David Reitter. 2022.
\newblock \href {https://doi.org/10.1162/tacl_a_00506} {Evaluating attribution
  in dialogue systems: The {BEGIN} benchmark}.
\newblock \emph{Transactions of the Association for Computational Linguistics},
  10:1066--1083.

\bibitem[{Elazar et~al.(2021)Elazar, Kassner, Ravfogel, Ravichander, Hovy,
  Sch{\"u}tze, and Goldberg}]{elazar-etal-2021-measuring}
Yanai Elazar, Nora Kassner, Shauli Ravfogel, Abhilasha Ravichander, Eduard
  Hovy, Hinrich Sch{\"u}tze, and Yoav Goldberg. 2021.
\newblock \href {https://doi.org/10.1162/tacl_a_00410} {Measuring and improving
  consistency in pretrained language models}.
\newblock \emph{Transactions of the Association for Computational Linguistics},
  9:1012--1031.

\bibitem[{Ganguli et~al.(2023)Ganguli, Askell, Schiefer, Liao,
  Luko{\v{s}}i{\=u}t{\.e}, Chen, Goldie, Mirhoseini, Olsson, Hernandez
  et~al.}]{ganguli2023capacity}
Deep Ganguli, Amanda Askell, Nicholas Schiefer, Thomas Liao, Kamil{\.e}
  Luko{\v{s}}i{\=u}t{\.e}, Anna Chen, Anna Goldie, Azalia Mirhoseini, Catherine
  Olsson, Danny Hernandez, et~al. 2023.
\newblock The capacity for moral self-correction in large language models.
\newblock \emph{arXiv preprint arXiv:2302.07459}.

\bibitem[{Gao et~al.(2022)Gao, Dai, Pasupat, Chen, Chaganty, Fan, Zhao, Lao,
  Lee, Juan et~al.}]{gao2022rarr}
Luyu Gao, Zhuyun Dai, Panupong Pasupat, Anthony Chen, Arun~Tejasvi Chaganty,
  Yicheng Fan, Vincent~Y Zhao, Ni~Lao, Hongrae Lee, Da-Cheng Juan, et~al. 2022.
\newblock Rarr: Researching and revising what language models say, using
  language models.
\newblock \emph{arXiv preprint arXiv:2210.08726}.

\bibitem[{Gekhman et~al.(2023)Gekhman, Herzig, Aharoni, Elkind, and
  Szpektor}]{gekhman2023trueteacher}
Zorik Gekhman, Jonathan Herzig, Roee Aharoni, Chen Elkind, and Idan Szpektor.
  2023.
\newblock Trueteacher: Learning factual consistency evaluation with large
  language models.
\newblock \emph{arXiv preprint arXiv:2305.11171}.

\bibitem[{Guo et~al.(2017)Guo, Pleiss, Sun, and Weinberger}]{pmlr-v70-guo17a}
Chuan Guo, Geoff Pleiss, Yu~Sun, and Kilian~Q. Weinberger. 2017.
\newblock \href {https://proceedings.mlr.press/v70/guo17a.html} {On calibration
  of modern neural networks}.
\newblock In \emph{Proceedings of the 34th International Conference on Machine
  Learning}, volume~70 of \emph{Proceedings of Machine Learning Research},
  pages 1321--1330. PMLR.

\bibitem[{Hao et~al.(2022)Hao, Tan, Tang, Zhang, Xing, and Hu}]{hao2022bertnet}
Shibo Hao, Bowen Tan, Kaiwen Tang, Hengzhe Zhang, Eric~P Xing, and Zhiting Hu.
  2022.
\newblock Bertnet: Harvesting knowledge graphs from pretrained language models.
\newblock \emph{arXiv preprint arXiv:2206.14268}.

\bibitem[{Hendrycks et~al.(2021)Hendrycks, Carlini, Schulman, and
  Steinhardt}]{hendrycks2021unsolved}
Dan Hendrycks, Nicholas Carlini, John Schulman, and Jacob Steinhardt. 2021.
\newblock Unsolved problems in ml safety.
\newblock \emph{arXiv preprint arXiv:2109.13916}.

\bibitem[{Honovich et~al.(2022)Honovich, Aharoni, Herzig, Taitelbaum,
  Kukliansy, Cohen, Scialom, Szpektor, Hassidim, and Matias}]{honovich2022true}
Or~Honovich, Roee Aharoni, Jonathan Herzig, Hagai Taitelbaum, Doron Kukliansy,
  Vered Cohen, Thomas Scialom, Idan Szpektor, Avinatan Hassidim, and Yossi
  Matias. 2022.
\newblock True: Re-evaluating factual consistency evaluation.
\newblock \emph{arXiv preprint arXiv:2204.04991}.

\bibitem[{Honovich et~al.(2021)Honovich, Choshen, Aharoni, Neeman, Szpektor,
  and Abend}]{honovich-etal-2021-q2}
Or~Honovich, Leshem Choshen, Roee Aharoni, Ella Neeman, Idan Szpektor, and Omri
  Abend. 2021.
\newblock \href {https://doi.org/10.18653/v1/2021.emnlp-main.619} {$q^{2}$:
  {E}valuating factual consistency in knowledge-grounded dialogues via question
  generation and question answering}.
\newblock In \emph{Proceedings of the 2021 Conference on Empirical Methods in
  Natural Language Processing}, pages 7856--7870, Online and Punta Cana,
  Dominican Republic. Association for Computational Linguistics.

\bibitem[{Huang et~al.(2022)Huang, Xia, Xiao, Chan, Liang, Florence, Zeng,
  Tompson, Mordatch, Chebotar et~al.}]{huang2022inner}
Wenlong Huang, Fei Xia, Ted Xiao, Harris Chan, Jacky Liang, Pete Florence, Andy
  Zeng, Jonathan Tompson, Igor Mordatch, Yevgen Chebotar, et~al. 2022.
\newblock Inner monologue: Embodied reasoning through planning with language
  models.
\newblock \emph{arXiv preprint arXiv:2207.05608}.

\bibitem[{Jiang et~al.(2021)Jiang, Araki, Ding, and
  Neubig}]{jiang-etal-2021-know}
Zhengbao Jiang, Jun Araki, Haibo Ding, and Graham Neubig. 2021.
\newblock \href {https://doi.org/10.1162/tacl_a_00407} {How can we know when
  language models know? on the calibration of language models for question
  answering}.
\newblock \emph{Transactions of the Association for Computational Linguistics},
  9:962--977.

\bibitem[{Jiang et~al.(2020)Jiang, Xu, Araki, and
  Neubig}]{jiang-etal-2020-know}
Zhengbao Jiang, Frank~F. Xu, Jun Araki, and Graham Neubig. 2020.
\newblock \href {https://doi.org/10.1162/tacl_a_00324} {How can we know what
  language models know?}
\newblock \emph{Transactions of the Association for Computational Linguistics},
  8:423--438.

\bibitem[{Joshi et~al.(2017)Joshi, Choi, Weld, and
  Zettlemoyer}]{joshi2017triviaqa}
Mandar Joshi, Eunsol Choi, Daniel~S Weld, and Luke Zettlemoyer. 2017.
\newblock Triviaqa: A large scale distantly supervised challenge dataset for
  reading comprehension.
\newblock \emph{arXiv preprint arXiv:1705.03551}.

\bibitem[{Jung et~al.(2022)Jung, Qin, Welleck, Brahman, Bhagavatula, Bras, and
  Choi}]{Jung2022MaieuticPL}
Jaehun Jung, Lianhui Qin, Sean Welleck, Faeze Brahman, Chandra Bhagavatula,
  Ronan~Le Bras, and Yejin Choi. 2022.
\newblock Maieutic prompting: Logically consistent reasoning with recursive
  explanations.
\newblock In \emph{Conference on Empirical Methods in Natural Language
  Processing}.

\bibitem[{Kadavath et~al.(2022)Kadavath, Conerly, Askell, Henighan, Drain,
  Perez, Schiefer, Dodds, DasSarma, Tran-Johnson et~al.}]{kadavath2022language}
Saurav Kadavath, Tom Conerly, Amanda Askell, Tom Henighan, Dawn Drain, Ethan
  Perez, Nicholas Schiefer, Zac~Hatfield Dodds, Nova DasSarma, Eli
  Tran-Johnson, et~al. 2022.
\newblock Language models (mostly) know what they know.
\newblock \emph{arXiv preprint arXiv:2207.05221}.

\bibitem[{Kamath et~al.(2020)Kamath, Jia, and Liang}]{kamath2020selective}
Amita Kamath, Robin Jia, and Percy Liang. 2020.
\newblock Selective question answering under domain shift.
\newblock \emph{arXiv preprint arXiv:2006.09462}.

\bibitem[{Kamoi et~al.(2023)Kamoi, Goyal, and
  Durrett}]{kamoi-etal-2023-shortcomings}
Ryo Kamoi, Tanya Goyal, and Greg Durrett. 2023.
\newblock \href {https://aclanthology.org/2023.eacl-main.11} {Shortcomings of
  question answering based factuality frameworks for error localization}.
\newblock In \emph{Proceedings of the 17th Conference of the European Chapter
  of the Association for Computational Linguistics}, pages 132--146, Dubrovnik,
  Croatia. Association for Computational Linguistics.

\bibitem[{Kuhn et~al.(2023)Kuhn, Gal, and Farquhar}]{kuhn2023semantic}
Lorenz Kuhn, Yarin Gal, and Sebastian Farquhar. 2023.
\newblock Semantic uncertainty: Linguistic invariances for uncertainty
  estimation in natural language generation.
\newblock \emph{arXiv preprint arXiv:2302.09664}.

\bibitem[{Kwiatkowski et~al.(2019)Kwiatkowski, Palomaki, Redfield, Collins,
  Parikh, Alberti, Epstein, Polosukhin, Devlin, Lee
  et~al.}]{kwiatkowski2019natural}
Tom Kwiatkowski, Jennimaria Palomaki, Olivia Redfield, Michael Collins, Ankur
  Parikh, Chris Alberti, Danielle Epstein, Illia Polosukhin, Jacob Devlin,
  Kenton Lee, et~al. 2019.
\newblock Natural questions: a benchmark for question answering research.
\newblock \emph{Transactions of the Association for Computational Linguistics},
  7:453--466.

\bibitem[{Lester et~al.(2021)Lester, Al-Rfou, and Constant}]{lester2021power}
Brian Lester, Rami Al-Rfou, and Noah Constant. 2021.
\newblock The power of scale for parameter-efficient prompt tuning.
\newblock \emph{arXiv preprint arXiv:2104.08691}.

\bibitem[{Lin et~al.(2022)Lin, Hilton, and Evans}]{lin2022teaching}
Stephanie Lin, Jacob Hilton, and Owain Evans. 2022.
\newblock Teaching models to express their uncertainty in words.
\newblock \emph{arXiv preprint arXiv:2205.14334}.

\bibitem[{Liu et~al.(2023{\natexlab{a}})Liu, Zhang, and
  Liang}]{liu2023evaluating}
Nelson~F Liu, Tianyi Zhang, and Percy Liang. 2023{\natexlab{a}}.
\newblock Evaluating verifiability in generative search engines.
\newblock \emph{arXiv preprint arXiv:2304.09848}.

\bibitem[{Liu et~al.(2023{\natexlab{b}})Liu, Yuan, Fu, Jiang, Hayashi, and
  Neubig}]{liu2023pre}
Pengfei Liu, Weizhe Yuan, Jinlan Fu, Zhengbao Jiang, Hiroaki Hayashi, and
  Graham Neubig. 2023{\natexlab{b}}.
\newblock Pre-train, prompt, and predict: A systematic survey of prompting
  methods in natural language processing.
\newblock \emph{ACM Computing Surveys}, 55(9):1--35.

\bibitem[{Long(2023)}]{long2023large}
Jieyi Long. 2023.
\newblock Large language model guided tree-of-thought.
\newblock \emph{arXiv preprint arXiv:2305.08291}.

\bibitem[{Mallen et~al.(2022)Mallen, Asai, Zhong, Das, Hajishirzi, and
  Khashabi}]{mallen2022not}
Alex Mallen, Akari Asai, Victor Zhong, Rajarshi Das, Hannaneh Hajishirzi, and
  Daniel Khashabi. 2022.
\newblock When not to trust language models: Investigating effectiveness and
  limitations of parametric and non-parametric memories.
\newblock \emph{arXiv preprint arXiv:2212.10511}.

\bibitem[{Maynez et~al.(2020)Maynez, Narayan, Bohnet, and
  McDonald}]{maynez-etal-2020-faithfulness}
Joshua Maynez, Shashi Narayan, Bernd Bohnet, and Ryan McDonald. 2020.
\newblock \href {https://doi.org/10.18653/v1/2020.acl-main.173} {On
  faithfulness and factuality in abstractive summarization}.
\newblock In \emph{Proceedings of the 58th Annual Meeting of the Association
  for Computational Linguistics}, pages 1906--1919, Online. Association for
  Computational Linguistics.

\bibitem[{Meng et~al.(2022)Meng, Bau, Andonian, and
  Belinkov}]{meng2022locating}
Kevin Meng, David Bau, Alex~J Andonian, and Yonatan Belinkov. 2022.
\newblock \href {https://openreview.net/forum?id=-h6WAS6eE4} {Locating and
  editing factual associations in {GPT}}.
\newblock In \emph{Advances in Neural Information Processing Systems}.

\bibitem[{Newman et~al.(2021)Newman, Choubey, and Rajani}]{newman2021p}
Benjamin Newman, Prafulla~Kumar Choubey, and Nazneen Rajani. 2021.
\newblock P-adapters: Robustly extracting factual information from language
  models with diverse prompts.
\newblock \emph{arXiv preprint arXiv:2110.07280}.

\bibitem[{Ouyang et~al.(2022)Ouyang, Wu, Jiang, Almeida, Wainwright, Mishkin,
  Zhang, Agarwal, Slama, Ray et~al.}]{ouyang2022training}
Long Ouyang, Jeffrey Wu, Xu~Jiang, Diogo Almeida, Carroll Wainwright, Pamela
  Mishkin, Chong Zhang, Sandhini Agarwal, Katarina Slama, Alex Ray, et~al.
  2022.
\newblock Training language models to follow instructions with human feedback.
\newblock \emph{Advances in Neural Information Processing Systems},
  35:27730--27744.

\bibitem[{Park et~al.(2022)Park, Popowski, Cai, Morris, Liang, and
  Bernstein}]{ParkPCMLB22}
Joon~Sung Park, Lindsay Popowski, Carrie~J. Cai, Meredith~Ringel Morris, Percy
  Liang, and Michael~S. Bernstein. 2022.
\newblock \href {https://doi.org/10.1145/3526113.3545616} {Social simulacra:
  Creating populated prototypes for social computing systems}.
\newblock In \emph{The 35th Annual {ACM} Symposium on User Interface Software
  and Technology, {UIST} 2022, Bend, OR, USA, 29 October 2022 - 2 November
  2022}, pages 74:1--74:18. {ACM}.

\bibitem[{Petroni et~al.(2019)Petroni, Rockt{\"a}schel, Lewis, Bakhtin, Wu,
  Miller, and Riedel}]{petroni2019language}
Fabio Petroni, Tim Rockt{\"a}schel, Patrick Lewis, Anton Bakhtin, Yuxiang Wu,
  Alexander~H Miller, and Sebastian Riedel. 2019.
\newblock Language models as knowledge bases?
\newblock \emph{arXiv preprint arXiv:1909.01066}.

\bibitem[{Press et~al.(2022)Press, Zhang, Min, Schmidt, Smith, and
  Lewis}]{press2022measuring}
Ofir Press, Muru Zhang, Sewon Min, Ludwig Schmidt, Noah~A Smith, and Mike
  Lewis. 2022.
\newblock Measuring and narrowing the compositionality gap in language models.
\newblock \emph{arXiv preprint arXiv:2210.03350}.

\bibitem[{Qin and Eisner(2021)}]{qin-eisner-2021-learning}
Guanghui Qin and Jason Eisner. 2021.
\newblock \href {https://doi.org/10.18653/v1/2021.naacl-main.410} {Learning how
  to ask: Querying {LM}s with mixtures of soft prompts}.
\newblock In \emph{Proceedings of the 2021 Conference of the North American
  Chapter of the Association for Computational Linguistics: Human Language
  Technologies}, pages 5203--5212, Online. Association for Computational
  Linguistics.

\bibitem[{Rashkin et~al.(2021)Rashkin, Nikolaev, Lamm, Aroyo, Collins, Das,
  Petrov, Tomar, Turc, and Reitter}]{rashkin2021measuring}
Hannah Rashkin, Vitaly Nikolaev, Matthew Lamm, Lora Aroyo, Michael Collins,
  Dipanjan Das, Slav Petrov, Gaurav~Singh Tomar, Iulia Turc, and David Reitter.
  2021.
\newblock Measuring attribution in natural language generation models.
\newblock \emph{arXiv preprint arXiv:2112.12870}.

\bibitem[{Schick et~al.(2023)Schick, Dwivedi-Yu, Dess{\`\i}, Raileanu, Lomeli,
  Zettlemoyer, Cancedda, and Scialom}]{schick2023toolformer}
Timo Schick, Jane Dwivedi-Yu, Roberto Dess{\`\i}, Roberta Raileanu, Maria
  Lomeli, Luke Zettlemoyer, Nicola Cancedda, and Thomas Scialom. 2023.
\newblock Toolformer: Language models can teach themselves to use tools.
\newblock \emph{arXiv preprint arXiv:2302.04761}.

\bibitem[{Schick et~al.(2022)Schick, Dwivedi-Yu, Jiang, Petroni, Lewis,
  Izacard, You, Nalmpantis, Grave, and Riedel}]{schick2022peer}
Timo Schick, Jane Dwivedi-Yu, Zhengbao Jiang, Fabio Petroni, Patrick Lewis,
  Gautier Izacard, Qingfei You, Christoforos Nalmpantis, Edouard Grave, and
  Sebastian Riedel. 2022.
\newblock Peer: A collaborative language model.
\newblock \emph{arXiv preprint arXiv:2208.11663}.

\bibitem[{Scialom et~al.(2021)Scialom, Dray, Lamprier, Piwowarski, Staiano,
  Wang, and Gallinari}]{scialom-etal-2021-questeval}
Thomas Scialom, Paul-Alexis Dray, Sylvain Lamprier, Benjamin Piwowarski, Jacopo
  Staiano, Alex Wang, and Patrick Gallinari. 2021.
\newblock \href {https://doi.org/10.18653/v1/2021.emnlp-main.529}
  {{Q}uest{E}val: Summarization asks for fact-based evaluation}.
\newblock In \emph{Proceedings of the 2021 Conference on Empirical Methods in
  Natural Language Processing}, pages 6594--6604, Online and Punta Cana,
  Dominican Republic. Association for Computational Linguistics.

\bibitem[{Shinn et~al.(2023)Shinn, Labash, and Gopinath}]{reflexion}
Noah Shinn, Beck Labash, and Ashwin Gopinath. 2023.
\newblock \href {https://doi.org/10.48550/arXiv.2303.11366} {Reflexion: an
  autonomous agent with dynamic memory and self-reflection}.
\newblock \emph{CoRR}, abs/2303.11366.

\bibitem[{Sun et~al.(2022)Sun, Yan, Abbeel, and Mordatch}]{sun2022quantifying}
Meiqi Sun, Wilson Yan, Pieter Abbeel, and Igor Mordatch. 2022.
\newblock \href {https://openreview.net/forum?id=LpBlkATV24M} {Quantifying
  uncertainty in foundation models via ensembles}.
\newblock In \emph{NeurIPS 2022 Workshop on Robustness in Sequence Modeling}.

\bibitem[{Tam et~al.(2022)Tam, Mascarenhas, Zhang, Kwan, Bansal, and
  Raffel}]{tam2022evaluating}
Derek Tam, Anisha Mascarenhas, Shiyue Zhang, Sarah Kwan, Mohit Bansal, and
  Colin Raffel. 2022.
\newblock Evaluating the factual consistency of large language models through
  summarization.
\newblock \emph{arXiv preprint arXiv:2211.08412}.

\bibitem[{Thorne et~al.(2018)Thorne, Vlachos, Cocarascu, Christodoulopoulos,
  and Mittal}]{thorne-etal-2018-fact}
James Thorne, Andreas Vlachos, Oana Cocarascu, Christos Christodoulopoulos, and
  Arpit Mittal. 2018.
\newblock \href {https://doi.org/10.18653/v1/W18-5501} {The fact extraction and
  {VER}ification ({FEVER}) shared task}.
\newblock In \emph{Proceedings of the First Workshop on Fact Extraction and
  {VER}ification ({FEVER})}, pages 1--9, Brussels, Belgium. Association for
  Computational Linguistics.

\bibitem[{Touvron et~al.(2023)Touvron, Lavril, Izacard, Martinet, Lachaux,
  Lacroix, Rozi{\`e}re, Goyal, Hambro, Azhar et~al.}]{touvron2023llama}
Hugo Touvron, Thibaut Lavril, Gautier Izacard, Xavier Martinet, Marie-Anne
  Lachaux, Timoth{\'e}e Lacroix, Baptiste Rozi{\`e}re, Naman Goyal, Eric
  Hambro, Faisal Azhar, et~al. 2023.
\newblock Llama: Open and efficient foundation language models.
\newblock \emph{arXiv preprint arXiv:2302.13971}.

\bibitem[{Varshney et~al.(2022)Varshney, Mishra, and
  Baral}]{varshney-etal-2022-investigating}
Neeraj Varshney, Swaroop Mishra, and Chitta Baral. 2022.
\newblock \href {https://doi.org/10.18653/v1/2022.findings-acl.158}
  {Investigating selective prediction approaches across several tasks in {IID},
  {OOD}, and adversarial settings}.
\newblock In \emph{Findings of the Association for Computational Linguistics:
  ACL 2022}, pages 1995--2002, Dublin, Ireland. Association for Computational
  Linguistics.

\bibitem[{Wang et~al.(2020)Wang, Cho, and Lewis}]{wang-etal-2020-asking}
Alex Wang, Kyunghyun Cho, and Mike Lewis. 2020.
\newblock \href {https://doi.org/10.18653/v1/2020.acl-main.450} {Asking and
  answering questions to evaluate the factual consistency of summaries}.
\newblock In \emph{Proceedings of the 58th Annual Meeting of the Association
  for Computational Linguistics}, pages 5008--5020, Online. Association for
  Computational Linguistics.

\bibitem[{Wang et~al.(2022)Wang, Wei, Schuurmans, Le, Chi, and
  Zhou}]{wang2022self}
Xuezhi Wang, Jason Wei, Dale Schuurmans, Quoc Le, Ed~Chi, and Denny Zhou. 2022.
\newblock Self-consistency improves chain of thought reasoning in language
  models.
\newblock \emph{arXiv preprint arXiv:2203.11171}.

\bibitem[{Wei et~al.(2022)Wei, Wang, Schuurmans, Bosma, Chi, Le, and
  Zhou}]{wei2022chain}
Jason Wei, Xuezhi Wang, Dale Schuurmans, Maarten Bosma, Ed~Chi, Quoc Le, and
  Denny Zhou. 2022.
\newblock Chain of thought prompting elicits reasoning in large language
  models.
\newblock \emph{arXiv preprint arXiv:2201.11903}.

\bibitem[{Welleck et~al.(2019)Welleck, Weston, Szlam, and
  Cho}]{welleck-etal-2019-dialogue}
Sean Welleck, Jason Weston, Arthur Szlam, and Kyunghyun Cho. 2019.
\newblock \href {https://doi.org/10.18653/v1/P19-1363} {Dialogue natural
  language inference}.
\newblock In \emph{Proceedings of the 57th Annual Meeting of the Association
  for Computational Linguistics}, pages 3731--3741, Florence, Italy.
  Association for Computational Linguistics.

\bibitem[{Wilder et~al.(1957)Wilder, Christie, Kurnitz, Trauner, Power,
  Dietrich, and Laughton}]{wilder1957witness}
Billy Wilder, Agatha Christie, Harry Kurnitz, Alexandre Trauner, Tyrone Power,
  Marl{\`e}ne Dietrich, and Charles Laughton. 1957.
\newblock \emph{Witness for the Prosecution}.
\newblock United Artists.

\bibitem[{Wright et~al.(2022)Wright, Wadden, Lo, Kuehl, Cohan, Augenstein, and
  Wang}]{wright-etal-2022-generating}
Dustin Wright, David Wadden, Kyle Lo, Bailey Kuehl, Arman Cohan, Isabelle
  Augenstein, and Lucy~Lu Wang. 2022.
\newblock \href {https://doi.org/10.18653/v1/2022.acl-long.175} {Generating
  scientific claims for zero-shot scientific fact checking}.
\newblock In \emph{Proceedings of the 60th Annual Meeting of the Association
  for Computational Linguistics (Volume 1: Long Papers)}, pages 2448--2460,
  Dublin, Ireland. Association for Computational Linguistics.

\bibitem[{Yoran et~al.(2023)Yoran, Wolfson, Bogin, Katz, Deutch, and
  Berant}]{yoran2023answering}
Ori Yoran, Tomer Wolfson, Ben Bogin, Uri Katz, Daniel Deutch, and Jonathan
  Berant. 2023.
\newblock Answering questions by meta-reasoning over multiple chains of
  thought.
\newblock \emph{arXiv preprint arXiv:2304.13007}.

\bibitem[{Yoshikawa and Okazaki(2023)}]{yoshikawa-okazaki-2023-selective}
Hiyori Yoshikawa and Naoaki Okazaki. 2023.
\newblock \href {https://aclanthology.org/2023.findings-eacl.150}
  {Selective-{LAMA}: Selective prediction for confidence-aware evaluation of
  language models}.
\newblock In \emph{Findings of the Association for Computational Linguistics:
  EACL 2023}, pages 2017--2028, Dubrovnik, Croatia. Association for
  Computational Linguistics.

\bibitem[{Zablotskaia et~al.(2023)Zablotskaia, Phan, Maynez, Narayan, Ren, and
  Liu}]{zablotskaia2023uncertainty}
Polina Zablotskaia, Du~Phan, Joshua Maynez, Shashi Narayan, Jie Ren, and
  Jeremiah Liu. 2023.
\newblock On uncertainty calibration and selective generation in probabilistic
  neural summarization: A benchmark study.
\newblock \emph{arXiv preprint arXiv:2304.08653}.

\bibitem[{Zhong et~al.(2021)Zhong, Friedman, and Chen}]{zhong2021factual}
Zexuan Zhong, Dan Friedman, and Danqi Chen. 2021.
\newblock Factual probing is [mask]: Learning vs. learning to recall.
\newblock In \emph{Proceedings of the 2021 Conference of the North American
  Chapter of the Association for Computational Linguistics: Human Language
  Technologies}, pages 5017--5033.

\bibitem[{Zhou et~al.(2022)Zhou, He, Ma, Berg-Kirkpatrick, and
  Neubig}]{zhou2022prompt}
Chunting Zhou, Junxian He, Xuezhe Ma, Taylor Berg-Kirkpatrick, and Graham
  Neubig. 2022.
\newblock Prompt consistency for zero-shot task generalization.
\newblock \emph{arXiv preprint arXiv:2205.00049}.

\end{thebibliography}
\bibliographystyle{acl_natbib}

\clearpage

\appendix

\section{Additional Evaluation}
\label{apx:more_evaluation}

We follow the same experimental setting as in \S\ref{sec:experiments}, but evaluate performance with respect to acceptance of claims rather than rejection. In addition, we introduce an ensemble \textbf{AYS + \crossexamination{}}; for a given claim, we first run the AYS method, and if the claim is rejected by this method we then apply \crossexamination{} (Majority) to obtain a final decision.

Tables~\ref{table:results_chatgpt_chatgpt_acceptance} and~\ref{table:results_gpt_gpt_acceptance} shows the evaluation results for the settings of \chatgpt{} vs. \chatgpt{} and \gptt{} vs. \gptt{}, respectively.

In terms of precision, our method outperforms the other baselines, often by a large gap (e.g., $81.6$ compared to $\leq60$ by baselines for \chatgpt{} on PopQA, and $68.7$ compared to $\leq52.4$ for \gptt{} on PopQA). Moreover, this is while maintaining a good recall performance with respect to the baselines, except for AYS that obtains the best recall scores. 
Importantly, surpass the baselines across all settings, overall obtaining the best F1 scores of $83.3-95.8$ for \chatgpt{} and $74-94.9$ for \gptt{} for factuality verification.
Moreover, unlike other methods, \crossexamination{} performs consistently well across datasets, including PopQA and NQ, which consider more challenging queries by real users or about rare entities that LMs struggle with \cite{mallen2022not}.

\begin{table*}[t]
\setlength\tabcolsep{4.0pt}
\footnotesize
\begin{center}
\begin{tabular}{l  ccc | ccc | ccc | ccc }
 & \multicolumn{3}{c}{LAMA} & \multicolumn{3}{c}{TriviaQA} &  \multicolumn{3}{c}{NQ} & \multicolumn{3}{c}{PopQA} \\ [0.5cm]
\multicolumn{1}{c}{} & \textbf{P}  & \textbf{R} & \textbf{F1}  & \textbf{P} & \textbf{R} & \textbf{F1}  & \textbf{P}  & \textbf{R} & \textbf{F1} & \textbf{P}  & \textbf{R} & \textbf{F1}  \\
\toprule
IDK                 & $88.6$    &$87.2$   &$87.9$ 
                    & $88.9$    &$79.3$   &$83.8$
                    & $72.0$    &$88.7$   &$79.5$
                    & $59.7$    &$68.7$   &$63.9$ \\
AYS                 & $84.8$    &$98.7$   &$91.2$   
                    & $80.3$    &$98.7$   &$88.5$                    
                    & $60.6$    &$95.6$   &$74.2$ 
                    & $53.5$    &$88.2$   &$66.6$ \\ \midrule
\textbf{\crossexamination{}}   & $93.3$    &$97.1$   &$95.2$
                            & $91.9$    &$95.6$   &$93.7$
                            & $81.2$    &$86.7$   &$83.9$ 
                            & $75.0$    &$82.0$   &$78.3$  \\ 
\textbf{\crossexamination{}} (Majority)   & $\mathbf{94.5}$    &$97.2$   &$\mathbf{95.8}$
                            & $\mathbf{94.4}$    &$95.6$   &$\mathbf{95.0}$
                            & $\mathbf{85.1}$    &$88.4$   &$\mathbf{86.7}$ 
                            & $\mathbf{81.7}$    &$85.0$   &$\mathbf{83.3}$  \\ 
\textbf{AYS + \crossexamination{}} (Ensemble)  & $83.3$    &$\mathbf{98.9}$   &$90.4$
                            & $78.9$    &$\mathbf{98.8}$   &$87.7$
                            & $58.9$    &$\mathbf{98.1}$   &$73.6$ 
                            & $49.0$    &$\mathbf{89.1}$   &$63.2$  \\
\bottomrule
\end{tabular}
\end{center}
\caption{Precision (P), Recall (R), and F1 scores of \chatgpt{} as \examinerLLM{} and \claimLLM{}.}
\label{table:results_chatgpt_chatgpt_acceptance}
\end{table*}

\begin{table*}[t]
\setlength\tabcolsep{4.0pt}
\footnotesize
\begin{center}
\begin{tabular}{l  ccc | ccc | ccc | ccc }
 & \multicolumn{3}{c}{LAMA} & \multicolumn{3}{c}{TriviaQA} &  \multicolumn{3}{c}{NQ} & \multicolumn{3}{c}{PopQA} \\ [0.5cm]
\multicolumn{1}{c}{} & \textbf{P}  & \textbf{R} & \textbf{F1}  & \textbf{P} & \textbf{R} & \textbf{F1}  & \textbf{P}  & \textbf{R} & \textbf{F1} & \textbf{P}  & \textbf{R} & \textbf{F1}  \\
\toprule
Confidence-Based    & $\mathbf{94.8}$    &$65.4$   &$77.4$ 
                    & $\mathbf{91.1}$    &$55.2$  &$68.7$ 
                    & $\mathbf{84.8}$    &$30.2$  &$44.5$ 
                    & $52.4$    &$42.6$   &$47.0$  \\
IDK                 & $85.4$    &$85.9$   &$85.6$ 
                    & $81.4$    &$82.7$   &$82.1$ 
                    & $56.6$    &$71.0$   &$63.0$ 
                    & $41.2$    &$44.9$   &$47.4$    \\
IC-IDK             & $85.2$    &$96.6$   &$90.5 $ 
                    & $84.1$    &$92.8$   &$88.3$ 
                    & $64.1$    &$75.4$   &$69.3$ 
                    & $51.8$    &$85.5$   &$64.5$   \\
AYS                 & $82.6$    &$98.5$   &$89.8$  
                    & $77.9$    &$98.3$   &$86.9$ 
                    & $54.1$    &$93.1$   &$68.4$ 
                    & $47.7$    &$90.1$   &$62.4$  \\ \midrule
\textbf{\crossexamination{}}   & $92.6$    &$95.2$   &$93.9$
                            & $88.5$    &$94.9$   &$91.6$
                            & $68.5$    &$72.2$   &$70.3$ 
                            & $64.4$    &$71.3$   &$67.7$  \\ 
\textbf{\crossexamination{}} (Majority)   & $94.5$    &$95.3$   &$\mathbf{94.9}$  
                            & $90.4$    &$94.9$  &$\mathbf{92.6}$  
                            & $77.6$    &$80.0$   &$\mathbf{78.8}$ 
                            & $\mathbf{68.7}$  & $80.2$  &$\textbf{74.0}$  \\
\textbf{AYS + \crossexamination{}} (Ensemble) & $80.5$    &$\mathbf{98.7}$   &$88.7$
                            & $76.0$    &$\mathbf{98.9}$   &$85.9$
                            & $52.8$    &$\mathbf{98.5}$   &$68.7$ 
                            & $46.6$    &$\textbf{91.9}$   &$61.8$   \\
\bottomrule
\end{tabular}
\end{center}
\caption{Precision (P), Recall (R), and F1 scores of \gptt{} as \examinerLLM{} and \claimLLM{}.}
\label{table:results_gpt_gpt_acceptance}
\end{table*}

\section{Falsehoods Data}
\label{apx:fake_facts}

To generate a wrong claim, given a query $Q$ for one of the QA datasets we use, we prompt our models the following way: in case $Q$ is in a question format, we use \texttt{``Please answer the following question with a wrong answer: <$Q$>''} and further request the LM to \texttt{``Please also phrase your answer as an argument.''}. 
in case $Q$ is in a sentence-completion format, we use \texttt{``Please complete the following sentence with a wrong answer: <$Q$>''} and further concatenate $Q$ with the model answer. Table~\ref{appx_table:false_claims} introduces a few examples of these, generated by \gptt{}.

\begin{table*}[t]
\setlength{\belowcaptionskip}{-10pt}
    \footnotesize
    \centering
    \setlength{\tabcolsep}{2.8pt}
    \begin{tabular}{lp{6.2cm}p{6.2cm}p{1.3cm}}
         & \textbf{False claim} & \textbf{True claim} & \textbf{Dataset}  \\ \midrule
         
        & \nl{Windows is an operating system developed by Apple.} & \nl{Windows is an operating system developed by Microsoft.} & LAMA \\ \midrule
        
        & \nl{The Hispaniolan lizard cuckoo (Coccyzus longirostris) is a species of cuckoo in the Cuculidae family.It is found in the Dominican Republic and Honduras.} & \nl{The Hispaniolan lizard cuckoo (Coccyzus longirostris) is a species of cuckoo in the Cuculidae family.It is found in the Dominican Republic and Haiti.} & LAMA \\ \midrule
        
         & \nl{The first modern electric battery was demonstrated by Thomas Edison, an American inventor.} & \nl{The first modern electric battery was demonstrated by Alessandro Volta.} & TriviaQA \\ \midrule
         
        & \nl{I believe that the actor who played Rockford's father, "Rocky," in the TV series, "The Rockford Files," was Tom Selleck.}  & \nl{The actor who played Rockford's father, "Rocky," in the TV series, "The Rockford Files," was Noah Beery Jr.} & TriviaQA \\ \midrule

        & \nl{The Taurus Mountains are located in the United States, specifically in the state of California.} & \nl{The Taurus Mountains are located in the southern Turkey} & NQ \\ \midrule

        & \nl{I heard that Taylor Swift is doing the 2018 Super Bowl Half Time Show.} & \nl{Justin Timberlake was the featured performer in the 2018 Super Bowl Half Time Show.} & NQ \\ \midrule

        & \nl{Red Velvet is a type of cake} & \nl{Red Velvet is a genre of music.} & PopQA \\ \midrule

        & \nl{The author of The Jacket is J.K. Rowling, because she is the author of the Harry Potter series.} & \nl{The author of The Jacker is Andrew Clements.} & PopQA \\ 
        
         \bottomrule
    \end{tabular}
    \caption{Example false claims generated by \gptt{} for our four QA datasets.}
    \label{appx_table:false_claims}
\end{table*}

\section{Example Cross-Examinations}
\label{apx:example_examinations}

Full cross-examination examples are provided in Tables~\ref{table:examination_example1},~\ref{table:examination_example2},~\ref{table:examination_example3},~\ref{table:examination_example4},~\ref{table:examination_example5},~\ref{table:examination_example6},~\ref{table:examination_example7},~\ref{table:examination_example8}.

\begin{table*}[t]
    \centering
    \footnotesize
    \begin{tabular}{|p{15.5cm}|}
    \toprule
\tcbox{\claimLLM{}}: \texttt{The song `I'm in the Mood for Dancing' was written by Irish songwriter and musician, Ben Findon.} \\ [0.5cm]
\tcbox{\examinerLLM{}}: \\
\texttt{1. Who is the artist that recorded the song \nl{I'm in the Mood for Dancing}?} \\
\texttt{2. When was the song \nl{I'm in the Mood for Dancing} first released?} \\
\texttt{3. Who are the listed songwriters for \nl{I'm in the Mood for Dancing}?} \\
\texttt{4. What is the nationality of songwriter Ben Findon?} \\
\texttt{5. Are there any credible sources or references that credit Ben Findon as the writer of \nl{I'm in the Mood for Dancing}?} \\ [0.5cm]
\tcbox{\claimLLM{}}:  \texttt{As an AI language model, I can answer the following questions based on available data:} \\
\texttt{1. The song \nl{I'm in the Mood for Dancing} was originally recorded by the Irish family band The Nolan Sisters, later known as The Nolans.} \\
\texttt{2. The song \nl{I'm in the Mood for Dancing} was first released in 1979, the year in which it became a major international hit.} \\
\texttt{3. The listed songwriters for \nl{I'm in the Mood for Dancing} are Ben Findon, Mike Myers, and Robert Puzey.} \\
\texttt{4. Ben Findon is a British songwriter.} \\
\texttt{5. Yes, there are credible sources that credit Ben Findon as the co-writer of \nl{I'm in the Mood for Dancing.} The song is registered with several music licensing agencies, such as ASCAP and BMI, which list Ben Findon as one of the song's writers. Additionally, the credits for the song are also listed in the liner notes of The Nolans' debut album of the same name.} \\ [0.5cm]
\tcbox{\examinerLLM{}}: \texttt{Can you provide any other sources or references that credit Ben Findon as the co-writer of \nl{I'm in the Mood for Dancing}?} \\ [0.5cm]
\tcbox{\claimLLM{}}: \texttt{As an AI language model, I can provide additional sources or references that credit Ben Findon as the co-writer of \nl{I'm in the Mood for Dancing.}} \\ 
\texttt{1. AllMusic, a music database and online resource, lists Ben Findon as one of the songwriters for \nl{I'm in the Mood for Dancing.}} \\
\texttt{2. BMI, a performing rights organization, lists Ben Findon as a co-writer of \nl{I'm in the Mood for Dancing} in their repertoire of musical works.} \\
\texttt{3. The official website of The Nolans also credits Ben Findon as one of the songwriters for \nl{I'm in the Mood for Dancing.}} \\
\texttt{4. The website of Discogs, a database and online marketplace for music, lists Ben Findon as a credited songwriter for the song.} \\
\texttt{5. The liner notes of The Nolans' album \nl{I'm in the Mood...} credit Ben Findon, Mike Myers, and Robert Puzey as the song's writers.} \\
\texttt{These sources all provide additional evidence that Ben Findon is a co-writer of \nl{I'm in the Mood for Dancing.}} \\ [0.5cm]
\tcbox{\examinerLLM{}}: \texttt{Based on the available information, the claim that \nl{The song 'I'm in the Mood for Dancing' was written by Irish songwriter and musician, Ben Findon} \textbf{is incorrect}. It appears that Ben Findon is only one of the co-writers of the song, alongside Mike Myers and Robert Puzey. 
The song was originally recorded by the Irish family band The Nolan Sisters, later known as The Nolans.} \\
\bottomrule
    \end{tabular}
    \caption{Example cross-examination with \chatgpt{} as \examinerLLM{} and  \claimLLM{}, for the claim \nl{The song `I'm in the Mood for Dancing' was written by Irish songwriter and musician, Ben Findon.} (which is correct). For brevity, prompts are not shown.}
    \label{table:examination_example1}
\end{table*}

\begin{table*}[t]
    \centering
    \footnotesize
    \begin{tabular}{|p{15.5cm}|}
    \toprule
\tcbox{\claimLLM{}}: \texttt{The producer of Precious was Lee Daniels.} \\ [0.5cm]
\tcbox{\examinerLLM{}}: \\
\texttt{1. Can you provide a list of the producers involved in the production of Precious?} \\
\texttt{2. Who was the director of Precious?} \\
\texttt{3. Did Lee Daniels win any awards for producing Precious?} \\
\texttt{4. What other movies has Lee Daniels produced?} \\
\texttt{5. Are there any interviews or articles where Lee Daniels specifically states that he was the producer of Precious?} \\ [0.5cm]
\tcbox{\claimLLM{}}:\\
\texttt{1. Sure, here is a list of producers involved in the production of ``Precious'': Lee Daniels, Gary Magness, Sarah Siegel-Magness, Oprah Winfrey, Tom Heller, and Tyler Perry.} \\ 
\texttt{2. The director of ``Precious'' is Lee Daniels.} \\ 
\texttt{3. Yes, Lee Daniels won several awards for producing ``Precious,'' including the Academy Award for Best Picture in 2010.} \\ 
\texttt{4. Lee Daniels has produced several other films throughout his career, including ``The Paperboy,'' ``Monster's Ball,'' ``The Woodsman,'' and ``The Butler.''} \\ 
\texttt{5. Yes, there are several interviews and articles in which Lee Daniels specifically states that he was one of the producers of ``Precious.'' In fact, he is often credited as one of the driving forces behind the film's success and is known for his efforts to bring the story to the screen.} \\ [0.5cm]
\tcbox{\examinerLLM{}}: \texttt{Based on the provided answers, the claim that the producer of ``Precious'' was Lee Daniels \textbf{is correct}.} \\
\bottomrule
    \end{tabular}
    \caption{Example cross-examination with \chatgpt{} as \examinerLLM{} and  \claimLLM{}, for the claim \nl{The producer of Precious was Lee Daniels.}  (which is correct). For brevity, prompts are not shown.}
    \label{table:examination_example2}
\end{table*}

\begin{table*}[t]
    \centering
    \footnotesize
    \begin{tabular}{|p{15.5cm}|}
    \toprule
\tcbox{\claimLLM{}}: \texttt{There are eight vertices (corners) on an octahedron} \\ [0.5cm]
\tcbox{\examinerLLM{}}: \\
\texttt{1. What is an octahedron?} \\
\texttt{2. How many faces does an octahedron have?} \\
\texttt{3. How many edges does an octahedron have?} \\
\texttt{4. How many vertices does an octahedron have? } \\
\texttt{5. Are all vertices of an octahedron located at the same distance from the center?} \\ [0.5cm]
\tcbox{\claimLLM{}}:\\
\texttt{1. An octahedron is a three-dimensional geometric shape that is made up of eight equal equilateral triangles. It is a polyhedron with eight faces.} \\ 
\texttt{2.  An octahedron has eight faces, each of which is an equilateral triangle.} \\ 
\texttt{3.  An octahedron has 12 edges that connect its eight vertices.} \\ 
\texttt{4. An octahedron has eight vertices, each of which is the point where three edges meet.} \\ 
\texttt{5. No, not all vertices of an octahedron are located at the same distance from the center. There are two types of vertices in an octahedron: those at the corners of the shape, which are equidistant from the center, and those located at the center of each face, which are further away from the center than the corner vertices.} \\ [0.5cm]
\tcbox{\examinerLLM{}}: \texttt{ Based on the answers, the claim that "there are eight vertices on an octahedron"  \textbf{is correct}.} \\
\bottomrule
    \end{tabular}
    \caption{Example cross-examination with \chatgpt{} as \examinerLLM{} and  \claimLLM{}, for the claim \nl{There are eight vertices (corners) on an octahedron.} (which is incorrect). For brevity, prompts are not shown.}
    \label{table:examination_example3}
\end{table*}

\begin{table*}[t]
    \centering
    \footnotesize
    \begin{tabular}{|p{15.5cm}|}
    \toprule
\tcbox{\claimLLM{}}: \texttt{Augustus was the first Roman Emperor to sport a beard.} \\ [0.5cm]
\tcbox{\examinerLLM{}}: \\
\texttt{1.Who was Augustus?} \\
\texttt{2. What is the definition of a Roman Emperor?} \\
\texttt{3. Was it common for Roman Emperors to wear beards?} \\
\texttt{4. Who were the previous Roman Emperors before Augustus? } \\
\texttt{5. Are there any historical records or paintings that depict Augustus with a beard?} \\ [0.5cm]
\tcbox{\claimLLM{}}:\\
\texttt{1. Augustus was the first Roman Emperor who ruled from 27 BCE to 14 CE. He was the successor of Julius Caesar and the founder of the Roman Empire.} \\ 
\texttt{2.  A Roman Emperor was a political leader who held supreme authority in the Roman Empire, including military, judicial, and legislative powers. The position was established after Julius Caesar's assassination, and Augustus was the first to hold the title officially.} \\ 
\texttt{3. No, it wasn't common for Roman Emperors to wear beards. In fact, most Roman Emperors were clean-shaven, and wearing a beard was perceived as a symbol of being barbaric or uncivilized.} \\ 
\texttt{4.The previous rulers before Augustus were known as Roman Republic's leaders, such as Julius Caesar, Pompey the Great, Marcus Licinius Crassus, and others.} \\ 
\texttt{5. There are no known historical records or paintings that depict Augustus with a beard. Most of the depictions show him with a clean-shaven face or a small mustache. Additionally, Augustus himself was known for promoting a clean-shaven look, and he even encouraged his advisors and officials to do the same.} \\ [0.5cm]
\tcbox{\examinerLLM{}}: \texttt{  Based on the information that has been gathered, the claim that Augustus was the first Roman Emperor to sport a beard  \textbf{is incorrect}. In fact, it was not common for Roman Emperors to wear beards, and Augustus was known for promoting a clean-shaven look. Additionally, there are no known historical records or paintings that depict Augustus with a beard.} \\
\bottomrule
    \end{tabular}
    \caption{Example cross-examination with \chatgpt{} as \examinerLLM{} and  \claimLLM{}, for the claim \nl{Augustus was the first Roman Emperor to sport a beard.}  (which is incorrect). For brevity, prompts are not shown.}
    \label{table:examination_example4}
\end{table*}

\begin{table*}[t]
    \centering
    \footnotesize
    \begin{tabular}{|p{15.5cm}|}
    \toprule
\tcbox{\claimLLM{}}: \texttt{The capital of Eritrea is Asmara.} \\ [0.5cm]
\tcbox{\examinerLLM{}}: \texttt{What is the official record of the capital of Eritrea according to the government?} \\ [0.5cm]
\tcbox{\claimLLM{}}: \texttt{The official record of the capital of Eritrea according to the government is Asmara.} \\ [0.5cm]
\tcbox{\examinerLLM{}}: \texttt{Does the government of Eritrea recognize any other cities as official capitals?} \\ [0.5cm]
\tcbox{\claimLLM{}}: \texttt{No, the government of Eritrea only recognizes Asmara as its official capital.} \\ [0.5cm]
\tcbox{\examinerLLM{}}: \texttt{Based on the answers to my questions, I conclude that the claim that the capital of Eritrea is Asmara \textbf{is correct}.} \\
\bottomrule
    \end{tabular}
    \caption{Example cross-examination with \gptt{} as \examinerLLM{} and  \claimLLM{}, for the claim \nl{The capital of Eritrea is Asmara} (which is correct). For brevity, prompts are not shown.}
    \label{table:examination_example5}
\end{table*}

\begin{table*}[t]
    \centering
    \footnotesize
    \begin{tabular}{|p{15.5cm}|}
    \toprule
\tcbox{\claimLLM{}}: \texttt{Croatian composer Igor Stravinsky was born in Zagreb.} \\ [0.5cm]
\tcbox{\examinerLLM{}}: \texttt{Where was Igor Stravinsky born?} \\ [0.5cm]
\tcbox{\claimLLM{}}: \texttt{Igor Stravinsky was born in Oranienbaum, Russia (now Lomonosov, Russia) on June 17, 1882.} \\ [0.5cm]
\tcbox{\examinerLLM{}}: \texttt{What was the nationality of Igor Stravinsky's parents?} \\ [0.5cm]
\tcbox{\claimLLM{}}: \texttt{Igor Stravinsky's parents were both Russian.} \\ [0.5cm]
\tcbox{\examinerLLM{}}: \texttt{Based on the information provided, the claim that Croatian composer Igor Stravinsky was born in Zagreb \textbf{is incorrect}.} \\
\bottomrule
    \end{tabular}
    \caption{Example cross-examination with \gptt{} as \examinerLLM{} and  \claimLLM{}, for the claim \nl{Croatian composer Igor Stravinsky was born in Zagreb}  (which is incorrect). For brevity, prompts are not shown.}
    \label{table:examination_example6}
\end{table*}

\begin{table*}[t]
    \centering
    \footnotesize
    \begin{tabular}{|p{15.5cm}|}
    \toprule
\tcbox{\claimLLM{}}: \texttt{Cristiano Ronaldo joined Manchester United in 2003.} \\ [0.5cm]
\tcbox{\examinerLLM{}}: \texttt{Which club did Cristiano Ronaldo join in 2003?} \\ [0.5cm]
\tcbox{\claimLLM{}}: \texttt{Cristiano Ronaldo joined Sporting CP in 2003.} \\ [0.5cm]
\tcbox{\examinerLLM{}}: \texttt{do you have any proof that Cristiano Ronaldo joined Sporting CP in 2003?} \\ [0.5cm]
\tcbox{\claimLLM{}}: \texttt{Yes, there is proof that Cristiano Ronaldo joined Sporting CP in 2003. Sporting CP has released official documents confirming the transfer, and there are also numerous news articles and interviews from the time that confirm the move.} \\ [0.5cm]
\tcbox{\examinerLLM{}}: \texttt{Based on the evidence provided, it is clear that the claim that Cristiano Ronaldo joined Manchester United in 2003 \textbf{is false}. He actually joined Sporting CP in 2003.} \\
\bottomrule
    \end{tabular}
    \caption{Example cross-examination with \gptt{} as \examinerLLM{} and  \claimLLM{}, for the claim \nl{Cristiano Ronaldo joined Manchester United in 2003}  (which is correct). For brevity, prompts are not shown.}
    \label{table:examination_example7}
\end{table*}

\begin{table*}[t]
    \centering
    \footnotesize
    \begin{tabular}{|p{15.5cm}|}
    \toprule
\tcbox{\claimLLM{}}: \texttt{The United States Capitol Building is topped by the Statue of Freedom.} \\ [0.5cm]
\tcbox{\examinerLLM{}}: \texttt{What is the actual figure that is located on top of the United States Capitol Building?} \\ [0.5cm]
\tcbox{\claimLLM{}}: \texttt{ The actual figure located on top of the United States Capitol Building is a bronze statue of the Roman goddess Freedom, which stands 19 feet 6 inches tall.} \\ [0.5cm]
\tcbox{\examinerLLM{}}: \texttt{ Is the figure known as the Statue of Freedom?} \\ [0.5cm]
\tcbox{\claimLLM{}}: \texttt{No, the figure is not known as the Statue of Freedom. It is known as the Statue of Liberty.} \\ [0.5cm]
\tcbox{\examinerLLM{}}: \texttt{Based on the answers to the questions, the claim that the United States Capitol Building is topped by the Statue of Freedom  \textbf{is incorrect}. The actual figure located on top of the United States Capitol Building is a bronze statue of the Roman goddess Freedom, which is not known as the Statue of Freedom.} \\
\bottomrule
    \end{tabular}
    \caption{Example cross-examination with \gptt{} as \examinerLLM{} and  \claimLLM{}, for the claim \nl{The United States Capitol Building is topped by the Statue of Freedom}  (which is correct). For brevity, prompts are not shown.}
    \label{table:examination_example8}
\end{table*}

\end{document}